\documentclass[sn-mathphys-num]{sn-jnl}

\usepackage{graphicx}
\usepackage{color, colortbl}
\definecolor{Gray}{gray}{0.95}
\usepackage[utf8]{inputenc}
\usepackage[T1]{fontenc}
\usepackage{url}
\usepackage{booktabs}
\usepackage{amsfonts}
\usepackage{nicefrac}
\usepackage{microtype}
\usepackage[table]{xcolor}
\usepackage{grffile} 
\usepackage{mathtools}
\usepackage{ragged2e}
\usepackage{amssymb}
\usepackage{caption}
\usepackage{algpseudocode}
\usepackage{makecell}
\usepackage{comment}
\usepackage{tikz}
\usepackage{lipsum,adjustbox}
\usepackage{algorithm,algpseudocode}
\usepackage{amsthm}
\usepackage{subfigure}
\usepackage{pdflscape}
\usepackage{amssymb}
\usepackage{pifont}
\usepackage{amsfonts}
\usepackage{tkz-euclide}
\usepackage{pythonhighlight}
\usepackage{multirow}
\usepackage{multicol}
\usepackage[english]{babel}
\usepackage[leftcaption]{sidecap}
\usepackage{amsmath}
\usepackage{paralist,pst-func, pst-plot, pst-math, pstricks-add,pgfplots}
\usepackage[edges]{forest}
\usepackage{arydshln}
\usepackage{mathtools}
\usepackage{array}
\usepackage{multirow}
\usepackage{booktabs}

\usepackage{rotating}
\usepackage{dcolumn} 
\usepackage{arydshln}
\newcolumntype{d}{D{.}{.}{2}}
\usepackage{fontawesome}
\usepackage{caption}
\usepackage{subcaption}
\DeclareCaptionLabelSeparator{none}{ }
\usepackage{tabularx}

\let\orcidlogo\relax
\usepackage{orcidlink}

\newcolumntype{b}{X}

\restylefloat{figure}
\restylefloat{table}

\makeatletter
\renewcommand{\fnum@figure}{\textbf{Fig. \thefigure}}
\makeatother

\usetikzlibrary{shadows.blur}
\usetikzlibrary{patterns,matrix,arrows}
\usetikzlibrary{positioning,shapes.symbols, arrows, calc}
\usetikzlibrary{decorations.pathreplacing,positioning, arrows.meta}

\newlength\gap
\usepackage{color, colortbl}
\definecolor{LightCyan}{rgb}{0.95,1,1}

\newcommand \FigPath[1]{#1}
\newcommand \FigRef[1]{Fig. \ref{#1}}
\newcommand \TableRef[1]{Table \ref{#1}}

\newcommand \EquationRef[1]{Eq. \ref{#1}}
\newcommand \SectionRef[1]{Section \ref{#1}}
\newcommand \AppendixRef[1]{Appendix \ref{#1}}
\graphicspath{{./}}

\newcommand\independent{\protect\mathpalette{\protect\independenT}{\perp}}
\def\independenT#1#2{\mathrel{\rlap{$#1#2$}\mkern2mu{#1#2}}}

\theoremstyle{thmstyleone}

\theoremstyle{thmstyletwo}

\theoremstyle{thmstylethree}

\newcommand{\cmark}{\textcolor{green!80!black}{\ding{51}}}
\newcommand{\xmark}{\textcolor{red!80!black}{\ding{55}}}

\begin{document}
	
\title[Article Title]{Out-of-distribution detection using normalizing flows on the data manifold}

\author*[1]{\fnm{Seyedeh Fatemeh} \sur{Razavi} \orcidlink{0000-0001-6764-4175}}\email{razavi\_f@ut.ac.ir}

\author[1]{\fnm{Mohammad Mahdi} \sur{Mehmanchi} \orcidlink{0000-0001-5268-8685}
}\email{mahdi.mehmanchi@ut.ac.ir}

\author[1]{\fnm{Reshad} \sur{Hosseini} \orcidlink{0000-0002-3669-760X}}\email{reshad.hosseini@ut.ac.ir}

\author[1]{\fnm{Mostafa} \sur{Tavassolipour} \orcidlink{0000-0003-0662-0115}}\email{tavassolipour@ut.ac.ir}

\affil[1]{\orgdiv{School of ECE, College of Engineering}, \orgname{University of Tehran},
\country{Iran}}


\abstract{
Using the intuition that out-of-distribution data have lower likelihoods, a common approach for out-of-distribution detection involves estimating the underlying data distribution.
Normalizing flows are likelihood-based generative models providing a tractable density estimation via dimension-preserving invertible transformations.
Conventional normalizing flows are prone to fail in out-of-distribution detection, because of the well-known curse of dimensionality problem of the likelihood-based models. To solve the problem of likelihood-based models, some works try to modify likelihood for example by incorporating a data complexity measure. We observed that these modifications are still insufficient.
According to the manifold hypothesis, real-world data often lie on a low-dimensional manifold. 
Therefore, we proceed by estimating the density on a low-dimensional manifold and calculating a distance from the manifold as a measure for out-of-distribution detection. We propose a powerful criterion that combines this measure with the modified likelihood measure based on data complexity. 
Extensive experimental results show that incorporating manifold learning while accounting for the estimation of data complexity
improves the out-of-distribution detection ability of normalizing flows.
This improvement is achieved without modifying the model structure or using auxiliary out-of-distribution data during training.
}


\maketitle

\section{Introduction}
\label{sec:introduction}
Out-Of-Distribution (OOD) detection classifies test data {as either in-distribution or out-of-distribution} \cite{yang2021generalized}.
Generative models that rely on the likelihood estimation are prominent candidates for detecting OOD data.
It is expected that these methods can assign low likelihood value{s} to OOD data.
However, high-dimensional likelihood-based generative models are {prone} to failure in OOD detection \citep{nalisnick2018deep}.

Normalizing Flows (NFs){,} as  likelihood-based generative models with tractable likelihood{s,} appear to be well-suited for addressing the OOD detection problem. However, their likelihood{s often fail to significantly distinguish} OOD data.
In fact, NFs tend to assign high likelihood to OOD data \citep{nalisnick2018deep, kirichenko2020normalizing}.
Therefore, it has been stated that they {primarily} learn local transformations {rather than capturing} semantic {information}.
This problem is fundamental because estimating the likelihood in high-dimensional spaces is challenging \citep{theis2015note}.
Using alternative {measures} instead of pure likelihood is a candidate solution followed by researchers such as likelihood ratio \citep{ren2019likelihood}, likelihood regret \citep{xiao2020likelihood}, and Input Complexity 
(IC) \citep{Serrà2020Input}.
To the best of our knowledge, the role of manifold learning {in addressing this} problem has not yet been {explored} for {single-image OOD detection using} NFs.

{A key} structural limitation of common NFs is that they cannot learn the embedded manifolds of the data.
Real{-world} data are {often} embedded in low-dimension{al} manifolds, and powerful generati{ve} models use this intuition \citep{kingma2013auto, goodfellow2014generative}.
NFs use bijective transformation{s}, and therefore they preserve dimensionality between the input and transformed spaces. One {might} think that this {limitation could be addressed} by using non-bijective transformation{s} that {map} the data space to a lower{-}dimensional {manifold. M}aximizing the likelihood of the data on {such a} manifold {could then be achieved.}
{However}, this optimization problem cannot be solved exactly{. A}s with some other well-known likelihood-based methods on the manifold{, these approaches only} approximate a lower-bound of the likelihood \citep{kingma2013auto}.

Recently, several researchers {have} proposed solutions to the problem of maximizing the likelihood of data on a manifold using NFs.
Some {approaches employ} injective transformation{s} in NFs \citep{brehmer2020flows, caterini2021rectangular, cunningham2020normalizing}.
{While} injective transformations {can address the issue, they} make the optimization computationally expensive.
{Brehmer and Cranmer} \cite{brehmer2020flows} {proposed} a two-step training procedure{:} first{,} the manifold is learned using {one} NF, {and} then the density is estimated using another NF. {Although t}his procedure simplifies training, it {may result in} poor density estimation.
Since learning only the manifold during the first training step{,} while ignoring the density estimation objective{,} raises {concerns} about acquiring a manifold that {may} introduce {complexities or even} irreparable challenges {for} density estimation {in} the second training step.
{Horvat and Pfister} \cite{horvat2021denoising} introduced a {single-step training} method that learns a low-dimensional manifold from NFs without {relying on} injective transformations.
{In this method}, the {manifold} learning {is} coupled with density estimation {in a single} NF{,} followed by another NF that {precisely} estimates the density on the learned manifold. This procedure {improves} the {integration} of manifold learning {and} density estimation, {setting} it {apart} from \cite{brehmer2020flows}.

\begin{figure}[t]
     \centering
     \subfigure[NLL on the manifold]{\includegraphics[width=0.31\textwidth, height=0.15\textwidth]{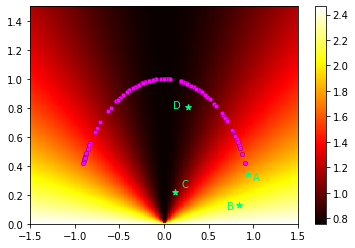}
     \label{fig:nll}}
     \subfigure[Reconstruction loss]{\includegraphics[width=0.31\textwidth, height=0.15\textwidth]{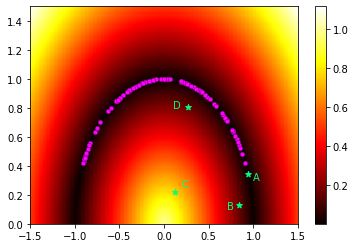}
     \label{fig:rec}}
     \subfigure[Proposed score]{\includegraphics[width=0.31\textwidth, height=0.15\textwidth]{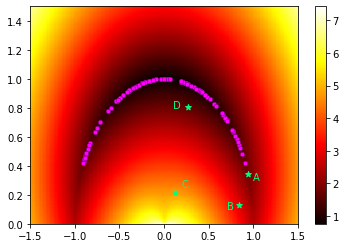}
     \label{fig:score}}
    \caption{A semicircle toy dataset illustrat{ing} the proposed score for OOD detection.
    This score combin{es} the NLL, which estimat{es} density, {with} a measurement of the distance to the manifold, {referred to as} the reconstruction loss.
    }
    \label{fig:idea}
\end{figure}

{
Building on previous methods, we propose an approach to address the limitations of NFs in estimating manifolds, particularly in high dimensions, and apply it to OOD detection.
To illustrate the idea, consider a toy dataset forming a semicircle, as shown in \FigRef{fig:idea}. The data lies on a one-dimensional manifold.
We transform each grid point by first projecting it onto the manifold and then mapping it back to the nearest point on the semicircle relative to its original position. Next, we calculate the reconstruction loss, which quantifies the distance between the original point and its mapped position using the Mean Squared Error (MSE). This process is illustrated in \FigRef{fig:rec}.
In addition, we estimate the density along the manifold using Negative Log-Likelihood (NLL), as shown in \FigRef{fig:nll}.

Relying only on NLL for OOD detection can lead to errors, as points with high likelihood but far from the manifold (e.g., point C in \FigRef{fig:nll}) may be misclassified as In-Distribution (ID). Similarly, using reconstruction loss alone may fail to distinguish between points in low-density and high-density regions (e.g., points B and D in \FigRef{fig:rec}, which have nearly identical reconstruction values but differ in density). To address this, we combine NLL and reconstruction loss into a unified OOD score that balances both terms. This approach ensures that points with balanced density and reconstruction loss are classified as ID (e.g., points A and D in \FigRef{fig:score}), while points in low-density regions or far from the manifold are classified as OOD (e.g., points B and C in \FigRef{fig:score}).
}

To jointly estimate the density and learn the manifold, the objective would be to optimize the likelihood {over} the on-manifold region while imposing penalties {for} the off-manifold region. To achieve this, a penalty function is employed, which encourages the inverse mapping to reconstruct the original data only on the on-manifold subspace.
We combine the achieved NLL with a positive factor of {the} distance from the manifold (reconstruction loss) to create an OOD detection score.
Summing different terms, such as NLL and the distance from the manifold, is not straightforward without {proper} standardization. To achieve {consistency} among these terms, we fit a distribution to the distance measurement{s}, such as a Gaussian distribution for MSE or a specialized density for other distance metrics.
The unique factor is determined based on the variance of this distribution.
We find this approach promising{,} as it enhances the performance of common NFs in OOD detection tasks without increasing architectural complexity or using auxiliary OOD training data.

In addition to addressing manifold learning, we also consider data complexity as a 
{critical} factor {influencing} likelihood estimation. Likelihood-based models are {typically} trained on specific data{sets} and {often} lack the desired {generalization}.
Consequently, relying only on likelihood scores for OOD detection may {be} insufficient.
{A} promising approach to {improve} likelihood estimation {is to use} a universal model that {operates} independent{ly} of the training data. {Such a} model can provide likelihood estimates while account{ing for} the complexities of the data.
In our work, we adopt {the} method {proposed} by \cite{Serrà2020Input}{, which refines} likelihood estimation by incorporating a data complexity estimator through the use of a universal lossless compressor.

Accordingly, our main contributions {can be} summarized as follows:
\begin{enumerate}
    \item We {investigate} the impact of incorporating manifold learning into NFs for OOD detection and demonstrate that {using} NFs with manifold learning can enhance OOD detection {in certain scenarios}.
    Furthermore, we {explore the integration of} data complexity as a test-time score into the proposed {detection framework}.
    
    \item While manifold learning with NFs algorithms typically employs the MSE penalty function, we {examine} the impact of using the Huber penalty function \citep{huber} during the manifold learning phase.
    {Our} find{ings suggest that the} Huber function {is} more suitable for OOD detection, primarily due to its {ability to adaptively} switch between linear and quadratic penalty {behaviors}.
    
    \item We {propose incorporating} a coefficient {for} the {reconstruction loss} in the {detection} score, proportional to the variance of the distribution fitted to the type of penalty function (Gaussian for MSE and 
    {a} density function for Huber). Our experiments demonstrate that this proportionality {is} independent of the manifold distance{, resulting in} a parameter-free method {that} effectively combin{es} the manifold and likelihood components {in a consistent manner}.
\end{enumerate}

The {structure of the} paper is as follows.
\SectionRef{sec:related_work} {provides a} review {of} recent methods related to manifold learning in NFs and OOD detection.
The {necessary} preliminaries are {outlined} in \SectionRef{sec:preliminaries}.
\SectionRef{sec:proposed_method} introduces the proposed method.
Experimental results are presented in \SectionRef{sec:results}.
Finally, the conclusion{s are} discussed in \SectionRef{sec:conclusion}.

\section{Related Work}
\label{sec:related_work}
Studies on manifold learning in NFs methods and OOD detection methods are provided in \SectionRef{sec:manifold_learning_in_nfs} and \SectionRef{sec:ood_detection}, respectively.

\subsection{Manifold learning using NFs}
\label{sec:manifold_learning_in_nfs}
This section provides a brief overview of the most relevant research about the manifold learning in NFs. In general, the methods are categorized into two classes: NFs on prescribed manifolds and NFs on learnable manifolds.
The first class focuses on learning flows on prescribed manifolds, while the second one includes learning the manifold, which aligns closely with our study.
We discuss the most related research in the second category in the next.

Recently, solutions to the problem of manifold learning and density estimation with NFs have been proposed \citep{kim2020softflow, kothari2021trumpets, cunningham2020normalizing,brehmer2020flows,caterini2021rectangular,horvat2021denoising,kalatzis2021multi,ross2021conformal, DBLP:conf/nips/GrcicGS21}.
In the following, we review the most related works to our research and highlight their key properties.
A proposed scientific path is about separating manifold learning and density estimation.
The pioneer of this study in NF is a method named $\mathcal{M}$-flow \citep{brehmer2020flows}.
In $\mathcal{M}$-flow, an injective NF transformation is used to transform the data into a manifold.
After that, another NF is used to estimate the density on the manifold. 
Manifold learning and density estimation are done separately to avoid the computation of Jacobian matrix induced by the injective transformation (two-phase training).
An extension of this study called multi-chart flow \citep{kalatzis2021multi}, which employs multiple mapping instead of one to find a manifold with multiple charts. It also suffers from two-phase training.
Another followed research of $\mathcal{M}$-flow, named Rectangular flow \citep{caterini2021rectangular},
overcomes the calculation of injective transformation by relying on automatic differentiation and linear algebra tricks.
A recent study, called Denoising NF (DNF) \citep{horvat2021denoising}, addresses the limitations of using injective transformations and separating model training by combining manifold learning and density estimation instead of treating them separately.
It splits the transformed space of an NF into two parts, noise-insensitive and noise-sensitive.
These parts are modeled by another NF and a low-variance Gaussian distribution, respectively. The noise is also added to the input training data.
Using this structure, two-phase training is no longer needed and the two NFs are trained simultaneously.

\subsection{Out-of-distribution detection}
\label{sec:ood_detection}

A brief overview of recent and prominent developments in OOD detection is provided in the following.
It should be noted that while there is a wide range of topics for this discussion \citep{yang2021generalized}, the main focus of the current research is OOD detection through density estimation.

\paragraph{Non-density-based methods}$ $

Contrary to the main focus of this paper, OOD detection methods are not limited to density-based models.
Using data labels can be a guide for this problem.
Common approaches include applying an appropriate threshold on a pre-trained classifier, reducing model overconfidence, altering the training approach, and using auxiliary OOD training data, as discussed by \cite{lee2017training,hendrycks2018deep,devries2018learning,liang2018enhancing}.

\cite{lee2017training} proposes a pre-training algorithm by using generated auxiliary OOD training data.
Since classifiers are not trained for the purpose of OOD detection, but instead to increase accuracy in inference time, a targeted pre-training approach can help the model enrich its features and decrease the model's confidence in OOD data.
Same to the mentioned research, one of the related researches (called outlier exposure) defines a specific loss function to discriminate between ID data and an auxiliary OOD data with a ranking loss to enhance the pre-training approach \citep{hendrycks2018deep}.
Confidence estimation along with the main objective function \citep{devries2018learning} or using a temperature-based Softmax function \citep{liang2018enhancing} are other solutions pursued in the literature.

\paragraph{Density-based methods}$ $

It seems likelihood-based generative models can be good candidates for OOD detection by assigning less probability to OOD data.
However, a frequent point about likelihood-based models such as Auto-Regressive (AR) models \citep{ren2019likelihood}, Variational Auto-Encoders (VAEs) \citep{xiao2020likelihood}, and NFs \citep{nalisnick2018deep,kirichenko2020normalizing} is that the likelihood is not a distinguishing score for OOD data \citep{UnderstandingBackground,UnderstandingFailure}.
The common weakness of all the mentioned models is that they also learn the background and irrelevant information. In other words, they equally attend to whole space.

Using the likelihood ratio instead of pure likelihood in AR models is a well-known solution that was proposed for the first time to solve the sequential genomics OOD detection \citep{ren2019likelihood}.
It considers a background/semantic decomposition of data.
The ratio score is obtained by dividing the likelihood value of the model by the likelihood value of a background model.
The background model is trained by perturbed data to learn the background information.
The defined ratio implicitly ignores irrelevant information.
Despite the success of this model in genomic OOD detection, it has not achieved state-of-the-art results on image datasets.
A remarkable solution for VAE is named likelihood regret \citep{xiao2020likelihood}.
This method is based on the principle that OOD data can drastically shift the likelihood after several fine-tuning steps. The authors employ a VAE to avoid overfitting in fine-tuning steps with freezing the decoder block.

According to the literature, NFs primarily learn local transformations rather than semantic features \citep{nalisnick2018deep, kirichenko2020normalizing}.
Therefore, despite having tractable likelihood, they have shortcomings in OOD detection.
Previously mentioned methods attempt to integrate high-level information in any way to address OOD issues.
For instance, some studies have demonstrated that training NFs on embedded data, as opposed to raw data, supports this claim \citep{kirichenko2020normalizing}.
Moreover, improving OOD detection can be achieved by redesigning the model's architecture. Using attention units \citep{kumar2021inflow} or deep residual flows \citep{zisselman2020deep} can be mentioned as pioneer works.
Similar to non-density-based methods, changes in training strategies (such as using information theory \citep{ardizzone2020training}) can be effective for OOD detection.
Employing a test-time OOD score based on input complexity is a valuable measurement \citep{Serrà2020Input}. It is considered as a likelihood ratio between the learned density and a universal lossless compressor.
Simple data can be encoded in fewer bits, making this ratio a means to estimate the actual number of bits needed.
A line of research overcomes this limitation by ensembling models \citep{choi2018waic}.
Surprisingly, this approach has been successful despite its over-parametrization.
Although individual density estimation models may not provide accurate diagnoses, this research demonstrates that the combination of such models, when considered in aggregate, can be successful against expectations.

Best of our knowledge, investigating the effect of manifold learning in NFs on the single image OOD detection has not been explored in the literature.
It is worth mentioning that the combination of density estimation and distance from the manifold as indicators in the context of anomaly detection in videos was explored in \cite{LandR}. A key distinction, from our perspective, lies in the calculation of likelihood. In their study, the likelihood is calculated in the transformed space, whereas in our approach, we perform manifold learning and density estimation simultaneously in the original space.
Besides, the two used terms (density and distance from the manifold) in their study are not homogeneous. To achieve a more meaningful combination, we have homogenized these terms by introducing the likelihood interpretation for the manifold distance.
Moreover, a medical image anomaly detection study adopts a similar approach to \cite{LandR} by employing evaluated likelihood in the transformed space for anomaly detection \citep{medicalanomaly}. It uses an encoder to transfer data into a latent space. Then, an NF is applied to transform the latent space into a Gaussian distribution, followed by bringing the transformed space back to the original using a decoder.
For training, its score combines the likelihood of the encoded space and a reconstruction loss. However, its anomaly detection score differs; it contains a linear combination of the structural similarity index measure between the original and reconstructed images, instead of a reconstruction loss.
As a result, its evaluation score diverges from the training score, the combination of its scores is not homogeneous, and the likelihood is not calculated in the original space.

\section{Preliminaries}
\label{sec:preliminaries}

This section introduces our notations and related preliminary concepts to make it easier for the reader to follow the subject.
The rest of this section is arranged as follows:
At first, the standard normalizing flow is discussed in \SectionRef{sec:nf}.
After that, a robust reconstruction loss function is presented in \SectionRef{sec:reconstruction}.

\subsection{Normalizing Flow}
\label{sec:nf}

There are a variety of well-known deep likelihood-based generative methods, like VAEs \citep{kingma2013auto}, NFs \citep{rezende2015variational}, AR models  \citep{murphy2022probabilistic}, energy base models \citep{murphy2022probabilistic}, and diffusion models \citep{NEURIPS2020_4c5bcfec}. 
But among the mentioned models,
only ARs and NFs can exactly compute the likelihood.
VAEs and diffusion models find a lower bound on the likelihood, while energy base models approximate it.
Sampling in common AR models is computationally expensive, due to sequential nature of these models.
Sampling in NFs is not sequential, but they have a structural limitation (dimension-preserving) that limits their applicability and generating power.

NF is a parametric diffeomorphism transformation,
$f_{\phi}:X \rightarrow Z$.
So, it is a two-side differentiable bijective transformation.
By choosing a random variable from a pre-defined (prior) distribution in $Z \in \mathbb{R}^D$ and transferring back it by $f_{\phi}^{-1}$, the distribution of data can be found by a change-of-variable formula like
\begin{equation}
    p_X(x) = p_Z(f_{\phi}(x)) \vert \det J_{f_\phi}(f_{\phi}(x)) \vert,
    \label{eq:nf}
\end{equation}
where $J_{f_\phi} \in \mathbb{R}^{D \times D}$ is the Jacobian matrix of the transformation $f_{\phi}$.
It is clear that the high computational complexity required for calculating the Jacobian determinant imposes an implicit constraint on NFs.
Moreover, the model parameters are estimated by minimizing NLL (equivalent to maximizing the log-likelihood) as
\begin{equation}
    \phi^{*} = \arg \underset{\phi}{\min} \big(- \log p_X(x)\big),
    \label{eq:nf_parameters}
\end{equation}
where $x=\{x_n\}_{n=1}^{n=N}$ is the available training data from the distribution $p_X$.

\subsection{Reconstruction loss functions}
\label{sec:reconstruction}
During the training of NFs on a manifold, all off-manifold data have a likelihood of 0.
In the literature, penalizing the off-manifold part by a quadratic reconstruction function, typically MSE, is a common approach to address this problem
\citep{brehmer2020flows,
caterini2021rectangular,
horvat2021denoising,
ross2021conformal}.
Quadratic functions exhibit sensitivity to far off-manifold data. A regularized reconstruction function combines linear and quadratic functions, dynamically switching based on the error and a threshold $\delta$. This adaptive switching leads less penalty for off-manifold data compared to on-manifold data.

\EquationRef{eq:huber} defines a switching function named the Huber function \citep{huber} with a strong background in robust statistics.
\begin{equation}
\label{eq:huber}
    H_\delta(e) = 
     \begin{cases}
      \frac{1}{2}e^2, & \text{if } e < \delta,
      \\
      \delta(e - \frac{1}{2} \delta), & \text{otherwise}.  \\ 
     \end{cases}
\end{equation}

\section{Proposed method}
\label{sec:proposed_method}

By considering a standard NF like $f_\phi$, the density of data is estimated by \EquationRef{eq:nf}.
It would be exciting if we could calculate the density of data on the manifold and measure its impact on OOD detection.
However, density estimation on manifolds is not an easy problem.
In this paper, we have inspired from existing methods of manifold learning in NFs ($\mathcal{M}$-flow \citep{brehmer2020flows} and DNF \citep{horvat2021denoising}) and propose a new one to detect OOD data.

From the manifold perspective, the transformed (or latent) space ($z$) can be disentangled into two sub-spaces, on-manifold ($u$), and off-manifold ($v$) space.
This is because placing all real-value data on a manifold is not necessarily a correct assumption, and some of the data points are generally outside the manifold.
The distribution of the transformed data is a joint distribution of these two parts like $p_Z(z) = p_Z(u,v)$.
By assuming independence of the two sub-spaces $u \independent v$, the distribution of the transformed space is decomposed into the product of two sub-space's distributions as $p_{Z}(z) = p_{U}(u) \times p_{V}(v)$, where $U: \mathcal{M}_M \subseteq \mathbb{R}^d$ and $V: \mathcal{M}_O \subseteq \mathbb{R}^{D-d}$ are the on-manifold and the off-manifold sub-spaces, respectively.

Our goal is to rearrange the transformed space of an NF in such a manner that a portion corresponds to the on-manifold part, while the other corresponds to the off-manifold part.
Let $f_\phi : \mathbb{R}^{D} \rightarrow \mathbb{R}^{D}$ be a standard NF. We denote the first $d$ dimensions of the output of $f_\phi$ as $u$ and the remaining dimensions as $v$ where $u$ corresponds to the on-manifold part of data and $v$ for the off-manifold part.
Formally,
$
z = f_\phi(x) = (z_1, z_2, ..., z_D),   u = (z_1, z_2, ..., z_d), v = (z_{d+1}, z_{d+2}, ..., z_D),
$
where $x$ is the input, and $z$ is the corresponding transformed data by the NF.
We choose $p_Z(z) = \mathcal{N}(0,I_D)$ and factorize $p_Z(z)$ such that $p_Z(z)=p_U(u) \times p_V(v)$.

It should be mentioned that we can use another NF $h_\theta$ to fit a more complex distribution on the on-manifold part 
($u$) rather than a Gaussian distribution.
This idea is inspired from DNF \citep{horvat2021denoising} and $\mathcal{M}$-flow \citep{brehmer2020flows} for density estimation.
Briefly, in this case, $p_{U}(u)$ is replaced by
$p_{U^\prime}(u^\prime)|\det G_{h_{\theta}}(u^\prime)|$,
where $u'$ is the transformed space by $h_{\theta}:\mathbb{R}^d \rightarrow \mathbb{R}^d$, and $p_{U^\prime}(u^\prime) = \mathcal{N}(0,I_d)$.
Moreover, DNF fits a tight Gaussian distribution $p_V \sim \mathcal{N}(0,\epsilon I_{D-d})$ on the off-manifold part based on the added noise value $\epsilon$ to the data.
However, the influence of the added noise in DNF was not reported to have a significant effect.
This is likely because the model is trained on real-valued data that have inherent quantization noise, making the addition of more noise unnecessary.
Consequently, we do not add noise to data in our framework, and so there are some minor differences with DNF.

The mentioned changes on standard NFs are not sufficient for manifold learning. It is necessary to penalize the off-manifold sub-space so that the model is able to reconstruct the data from the on-manifold sub-space.
Accordingly, we have a constrained optimization problem on NLL as
\begin{equation}
    \underset{\phi}{\min} \;\textrm{NLL}(x; \phi) \;\textrm{s.t.} \; \mathcal{C}(x,\tilde{x}) \le \tau,
    \label{eq:constrained_likelihood}
\end{equation}
where
$x = (x_1, x_2, ..., x_D)$,
$\tilde{x}
=
(\tilde{x}_1, \tilde{x}_2, ..., \tilde{x}_D)
=
f_\phi^{-1}(\text{proj}(f_\phi(x)))$,
$\mathcal{C}$, and
$\tau$
are the input data, corresponding reconstruction (computed through the inverse of normalizing flow $f_\phi$), the penalty function, and the penalization threshold, respectively.
Moreover, $\text{proj}(f_\phi(x))$ is the first $d$ components of $f_\phi(x)$ corresponding to the data manifold, padded with zero as $\text{proj}(f_\phi(x)) = (u, \vec{0}_{D-d})$.

Existing methods such as $\mathcal{M}$-flow \citep{brehmer2020flows}, DNF \citep{horvat2021denoising}, and Rectangular flow \citep{caterini2021rectangular} penalize the off-manifold with a quadratic constraint.
An important characteristic of a quadratic function is its sensitivity to outliers, that can be interpreted as off-manifold parts of the data.
One aspect of our contribution involves
using a switching penalization function, containing both linear and quadratic forms, which is determined based on the distance to the manifold.
The switching is modeled with a Huber function (Eq. \eqref{eq:huber}).
In the proposed method, an element-wise Huber function with threshold $\delta$, denoted $H_{\delta}$, is applied to the difference between the input data ($x$) and the reconstructed one ($\tilde{x}$). The reconstructed data is a function of on-manifold part ($\tilde{x} = f_\phi^{-1}(\text{proj}(f_\phi(x)))$) rather than joint on-manifold and off-manifold parts.
Therefore, the penalization term is computed by averaging an element-wise function like
\begin{equation}
    \mathcal{C}(x,\tilde{x}; \delta) = \frac{1}{D} \sum_{i=1}^{D} H_\delta(|x_i - \tilde{x}_i|).
\end{equation}

Inspired by the Lagrange multipliers method, the proposed constrained optimization problem can be converted to its unconstrained equivalent one

\begin{equation}
\ell(x) = -\log P_u(u) - \log P_v(v) + \log|\det J_{f_\phi}(x)| + \lambda \mathcal{C}(x,\tilde{x}; \delta),
\label{eq:loss}
\end{equation}
where $\lambda$ is the penalization term hyper-parameter, and the first three terms are the objective of a standard NF (called NLL).
The proposed approach not only facilitates the estimation of manifold density for NFs but also addresses the limitations of relying only on likelihood or reconstruction for OOD detection, as extensively explained in \SectionRef{sec:introduction} (see \FigRef{fig:idea}).  In other words, high likelihood or low reconstruction error alone are not reliable indicators for identifying ID data. However, by combining these two factors using the introduced cost function (\EquationRef{eq:loss}), we not only estimate the manifold using NFs but also introduce an appropriate indicator for OOD detection that is validated through comprehensive experiments.

As is evident, \EquationRef{eq:loss} incorporates a hyper-parameter both during training and scoring for OOD detection.
During training, a balance between the two terms (NLL and penalization) is maintained by setting $\lambda=1$.
However, during inference, an effort is made to employ a free parameterization approach to standardize the terms, enhancing efficiency.
To achieve this goal, the penalization term, i.e. 
$\mathcal{C}(x, \tilde{x}; \delta)$ is scaled by a coefficient determined by the associated distribution's variance. This adjustment applies a likelihood-based concept to this term, making the summation more homogeneous with the first term (NLL).
It is worth mentioning that when the penalty function is MSE, the distribution of the data is aligned with a Gaussian distribution, $\mathcal{N}(0, \sigma_{\text{mse}})$. In the case of Huber penalty function, the NLL of the Huber density function is as

\begin{equation}
\text{NLL}_{\text{Huber}}
=
- N \log (C_{\delta,k})
+ \frac{\sum_{n=1}^{n=N} H_{\delta}(|x_n-\tilde{x}_n|)}{k^2},
\label{eq:huber_density_original}
\end{equation}

where
$C_{\delta,k} = \Big(
	\frac{2k^2}{\delta}
	\exp(\frac{-\delta^2}{2k^2})
	+
	\sqrt{2\pi}k
	\big(
	2\Phi(\frac{\delta}{k}) - 1
	\big) \Big)^{-1}$
and $k^2$ is the variance of the Huber density function.
More detailed information about \EquationRef{eq:huber_density_original} is provided in \AppendixRef{sec:appendix}.
Briefly, the parameter $k$ is determined using Newton's optimization method to minimize \EquationRef{eq:huber_density_original}, given its lack of a closed-form solution. This process yields the optimal model parameter, which, in turn, guides the selection of the $\lambda$ coefficient as a fraction of the optimal $k$ value, typically set as $\lambda=\frac{C}{k^2}$.
In the case of MSE penalization, $\lambda$ is set to $\frac{C}{\sigma_{\text{mse}}}$.
Through experiments, we demonstrate that $C$ is not sensitive to errors in the both penalization types, allowing us to create a parameter-free method.

In addition to the previous discussion on manifold learning, we also address data complexity as an additional factor to enhance likelihood estimation and, subsequently, the proposed score.
To apply this concept, we rely on the work of \cite{Serrà2020Input} (referred to as IC), which re-estimates likelihood using a universal lossless compressor to modify the estimated likelihood provided by the models during test-time as
\begin{equation}
\ell_{\text{IC}}(x) = \text{NLL}(x) - \frac{|\mathbb{C}(x)|}{D},
\label{eq:IC_loss}
\end{equation}
where the second term signifies the number of bits generated through a data compression method (named $\mathbb{C}$), normalized with the data dimension $D$.

All in all, our final proposed score for OOD detection is
\begin{equation}
\ell(x) = -\log P_u(u) - \log P_v(v) + \log|\det J_{f_\phi}(x)| + \lambda \mathcal{C}(x,\tilde{x}; \delta) - \frac{|\mathbb{C}(x)|}{D}.
\label{eq:final_loss}
\end{equation}
We conduct numerous experiments to assess the impact of terms or penalty functions, as detailed in the next section.

\section{Results}
\label{sec:results}

The primary goal of the experiments is to explore the significance of manifold learning and data complexity in NFs for enhancing OOD detection.
Moreover, the performance of the proposed method is also evaluated in terms of image generation because it is fundamentally a generative model.
As previously mentioned, $\mathcal{M}$-flow \citep{brehmer2020flows} and DNF \citep{horvat2021denoising} are existing methods that closely resemble our approach for manifold learning in NFs.
Our proposed OOD detection score, combines the precise likelihood value with the reconstruction loss.
However, $\mathcal{M}$-flow cannot effectively achieve high-dimensional likelihood due to using injective transformations.
Therefore, our focus is on comparing our approach with the DNF framework in terms of image generation performance. Subsequently, we compare our OOD detection results with some other state-of-the-art likelihood-based OOD detection methods such as likelihood ratio \citep{ren2019likelihood} and IC \citep{Serrà2020Input}.
{
It is worth noting that, while diffusion models are effective in generation, they are typically categorized under reconstruction-based methods rather than likelihood-based methods in OOD detection tasks \citep{DBLP:conf/cvpr/GrahamPTNOC23}. Therefore, we compare our approach with the likelihood-based category, as diffusion models cannot achieve exact likelihood.
}

\subsection{Experimental setting}
The used dataset and experimental setting are described in \SectionRef{sec:dataset} and \SectionRef{sec:architecture}, respectively.

\subsubsection{Dataset}
\label{sec:dataset}

Six datasets used in the current study are introduced in \TableRef{tab:dataset}.
We could increase the number of layers and parameters in proportion to the original image's dimensions. Still, due to resource allocation limitations, we resize the color images to $32 \times 32$ in our experiments.

\begin{table}[thbp]
    \caption{The information of the used datasets in the paper.}
    \label{tab:dataset}
    \centering
    \begin{tabular}{p{0.3\textwidth}p{0.65\textwidth}}
    \toprule
    
    Data & Description \\
    
    \cmidrule{1-2}\morecmidrules\cmidrule{1-2}

    MNIST \citep{mnist} & Containing a set of $28 \times 28$ gray-scale hand-written digit images in 10 classes (50k training images, 10k test images) \\

    \hdashline
    
    FMNIST \citep{xiao2017fashionmnist} & Including a set of $28 \times 28$ gray-scale fashion products images from 10 different categories (60k training images, 10k test images).
    The dataset's original title is Fashion-MNIST, which we abbreviate as FMNIST. \\

	\hdashline
    
    SVHN \citep{svhn} & Consisting of $32 \times 32$ RGB house number plates (about 73k trainig images and 26k test images)\\

	\hdashline
    
    CIFAR10 \citep{cifar10} & Containing 60k $32 \times 32$ RGB images of 10 categories (50k training images, 10k test images)\\

    \hdashline
    
    CelebA \citep{liu2015faceattributes}
    & Consisting about 200k face images of 10,177 persons (180k training images, 20k test images). A resized version ($32 \times 32$) is used in this paper.  \\

    \hdashline
    
    LSUN \citep{yu2015lsun}
    & 
    Including a variety of large-scale images categorized into 10 classes.
    In this study, we only used bedroom class images. This class comprises more than 3 million images in the original dataset, but we use only
    about 300k
    $32 \times 32$ resized version
    of them.
    \\


    \bottomrule
\end{tabular}
\end{table}

\subsubsection{Architecture}
\label{sec:architecture}

For two-blocks models such as standard DNF or architecture searching experiments, we use the Glow model \citep{kingma2018glow} for $f_\phi$ and the RealNVP model \citep{dinh2016density} for $h_\theta$ explained in Section \ref{sec:proposed_method}. Meanwhile, for one-block models, we use either one of the models.
Our Glow model has three blocks, each with 32 steps of the flow.
Each step consists of an actnorm layer, an invertible $1 \times 1$ convolution layer, and an affine coupling layer (see \cite{kingma2018glow} for the details of each layer). There is also a squeeze layer before each step and a split layer after each step.
The RealNVP model comprises six affine coupling layers and six masking layers before each affine coupling layer.
While DNF has chosen a specific architecture in its experiments, the employed architectures in this paper are based on two well-known NFs, Glow and RealNVP.
Our main goal is to show the importance of manifold learning in NFs for OOD detection, regardless of the specific architecture.

\subsubsection{Training}
\label{sec:training}
All implementations are done in the PyTorch\footnote{https://pytorch.org/}(version 1.10.0+cu102) framework. We use Adam \citep{adam} as our optimization algorithm with a learning rate of 1e-5 and a batch size of 64.
All training experiments were done on a single GPU (GTX 1080 Ti), with Cuda version 11.4 and 11019GB memory.
The implementation code for the proposed method is available at the following link: \url{http://visionlab.ut.ac.ir/resources/ood.zip}.

\subsection{Architecture searching}
\label{sec:architecture_searching}

In this section, we consider four structures: DNF, MSE-based off-manifold penalization with two blocks (called Proposed-M), Huber-based off-manifold penalization with two blocks (called Proposed-H), and  Huber-based off-manifold penalization with one block (called Proposed-D).
In the first three, two NFs (named $f_\phi$ and $h_\theta$, Glow followed by RealNVP) are used, and in the last structure, a single NF (named $f_\phi$) is used.
Briefly, we employ two blocks together as an NF to be more comparable in structure with DNF.
Therefore, the two-blocks method means using a dimension preserving NF (RealNVP in our setting) to estimate the density of the manifold part (same as DNF) instead of the Gaussian distribution.
We choose all DNF backbone models the same as ours in all experiments for a fair comparison.

We only present the results of CelebA here and
the rest are available in the supplementary files.
Based on the reported results for CelebA dataset by \cite{horvat2021denoising} (DNF), we consider $d=|\mathcal{M}_M| \sim 500$ for architecture searching.
It is worth noting that, we assume prior knowledge of the manifold's intrinsic dimension from existing works in the field. Our goal is not to determine the intrinsic dimension of the manifold. Readers seeking dimensionality determination can refer to \cite{NEURIPS2022_4f918fa3} and \cite{DBLP:conf/iclr/PopeZAGG21}.

\begin{figure}[thbp]
\centering
\subfigure[DNF]{\includegraphics[width=0.24\textwidth]{\FigPath{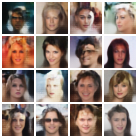}}}
\subfigure[Proposed-M]{\includegraphics[width=0.24\textwidth]{\FigPath{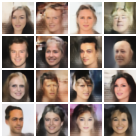}}}
\subfigure[Proposed-H]{\includegraphics[width=0.24\textwidth]{\FigPath{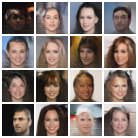}}}
\subfigure[Proposed-D]{\includegraphics[width=0.24\textwidth]{\FigPath{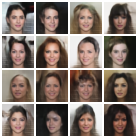}}}
\caption{Generated CelebA images corresponding to experiments in \TableRef{tab:table_face_manifold_CelebA}.}
\label{fig:figure_face_manifold_CelebA}
\end{figure}

The results of experiment on CelebA dataset are shown in \TableRef{tab:table_face_manifold_CelebA} and \FigRef{fig:figure_face_manifold_CelebA}.
The main goal of the designed experiment is to find an appropriate architecture for generating data from the manifold as long as it does not decrease the likelihood of data.
The reported results in \TableRef{tab:table_face_manifold_CelebA} confirm that our proposed methods
(Proposed-H,
Proposed-M, and
Proposed-D)
outperform DNF in terms of Bit-Per-Dim (BPD) (calculated by $\frac{\text{NLL}}{D \times \log2}$), whereas their generated images look visually similar, according to \FigRef{fig:figure_face_manifold_CelebA}.
An important aspect of DNF is its reliance on incorporating noise during training. It is noteworthy that the reported inference BPD for the DNF is computed using noiseless data, and making it an approximation.
It is worth mentioning that best generated images in \cite{horvat2021denoising} (DNF) are based on a well-defined StyleGAN manifold \citep{DBLP:conf/cvpr/KarrasLAHLA20}, not on raw data.
Additionally, their reported generated images on the CelebA dataset are not visually perfect.

\begin{table}[thbp]
\setlength{\tabcolsep}{0.5cm}
\caption{The best MSE/BPD scores for the CelebA data when $\mathcal{M} \subset \mathbb{R}^{500}$.
Dimension changing of the one-block (Proposed-D) and two-block (DNF, Proposed-M, Proposed-H) methods are $\mathbb{R}^{3072} \rightarrow \mathbb{R}^{500}$ and $\mathbb{R}^{3072} \rightarrow \mathbb{R}^{500} \rightarrow \mathbb{R}^{500}$.
\#Params means the number of trainable parameters.}
\label{tab:table_face_manifold_CelebA}
\centering
\begin{tabular}{p{0.05\textwidth}p{0.16\textwidth}p{0.16\textwidth}p{0.13\textwidth}p{0.13\textwidth}}
	
    \toprule	
    
    \scriptsize Criterion
    & \scriptsize DNF
    & \scriptsize \makecell{Proposed-M \\ \small (two-block)}
    & \scriptsize \makecell{Proposed-H \\ \small (two-block)}
    & \scriptsize \makecell{Proposed-D \\ \small (one-block)}
    \\
		
    \cmidrule{1-5}\morecmidrules\cmidrule{1-5}
    
    \scriptsize MSE & \scriptsize 0.004 $\pm$ 0.0002 & \scriptsize 0.005 $\pm$  0.0003 & \scriptsize 0.01 $\pm$ 0.001 &
    \scriptsize 0.02 $\pm$ 0.001 \\
    
    \scriptsize BPD & \small 3.98 $\pm$ 0.03 & \small 3.52 $\pm$ 0.04 & \small 3.51 $\pm$ 0.04 & \small 3.52 $\pm$ 0.04 \\

    \scriptsize \#Params & \small $\approx$ 59M & \small $\approx$ 59M & \small $\approx$ 59M & $\approx$ 44M \\

    \bottomrule

\end{tabular}
\end{table}

Since the Huber function applies a linear function to penalize the off-manifold pixels rather than a quadratic one, it is normal that the MSE would be higher in the models that were penalized with the Huber function.
In addition to the aforementioned factors, the integration of reconstruction loss and BPD value is a crucial component in our study. Although the reconstruction loss may slightly lag behind the DNF method in \TableRef{tab:table_face_manifold_CelebA}, our approach does not just depend on reconstruction or BPD (as scaled NLL) for OOD detection.
Based on the evaluation of the results in the latter three columns, we can confidently assert that the proposed method offers a promising advantage in this regard.
Another strength is that the Proposed-D method achieves same images (in terms of visual quality in \FigRef{fig:figure_face_manifold_CelebA}) to DNF and other two-blocks proposed methods, with fewer parameters while maintaining the likelihood.
Considering the outcomes, the proposed-D model is chosen for further experiments in the following. To summarize, we use the term 'proposed' to denote the model referred to as 'proposed-D'.

By choosing Proposed-D as the base model, more experiments are presented here to evaluate the performance of the model in terms of image generation and manifold learning on ID/OOD data.
The order of experiments is in such a way that the ability of the models is measured for different ID/OOD data and different manifold dimensions $(10, 50, 100, 500, 1000)$.
The results for CelebA test data are presented here. The results of other datasets are available in the supplementary file.

\FigRef{fig:figure_CelebA_generation_reconstruction} contains the generated and reconstructed data (learned manifold) for a model trained on CelebA.
In low-dimensional manifolds, images generated by the model are limited to displaying the main manifold of the face and the background details are not generated. However, as the manifold dimensions increase, the inclusion of less frequent details (mostly related to the background) in the generated image is observed.
If the data manifold has simplistic visual attributes (e.g., as observed in the SVHN dataset results presented in the supplementary file), the likelihood can be better fit to the data, and thus sharpened images can be generated as manifold dimension increases.
This trend is also observed in more complex datasets such as CIFAR10 (as noted in the supplementary file) when the likelihood is fitted using various dimensions of the manifold.

The second row of \FigRef{fig:figure_CelebA_generation_reconstruction} contains the reconstruction of CelebA test data on the model trained with CelebA for the corresponding manifold dimension. The reconstruction ability is improved by increasing the dimension.
As it is clear, the model's invertibility power will appear by increasing manifold dimension, the same as standard NF.
The model trained on CelebA not only excels at generating and reconstructing ID data but also demonstrates great performance in discriminating OOD data.
In case of reconstructiong OOD data in third and forth rows of \FigRef{fig:figure_CelebA_generation_reconstruction},
especially from low dimensions, it cannot reconstruct OOD data well.
In other words, learning low dimension manifold for ID data leads to focusing on ID specific features instead of global features.
It is worth noting that our point is about manifold learning while in case of dimension preserving (like standard NFs), the reconstruction is done without error.

\begin{figure}[thbp]
\centering

\subfigure[$\mathcal{M}_G \subset \mathbb{R}^{10}$]{\includegraphics[width=0.18\textwidth]{\FigPath{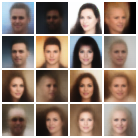}}}
\subfigure[$\mathcal{M}_G \subset \mathbb{R}^{50}$]{\includegraphics[width=0.18\textwidth]{\FigPath{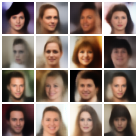}}}
\subfigure[$\mathcal{M}_G \subset \mathbb{R}^{100}$]{\includegraphics[width=0.18\textwidth]{\FigPath{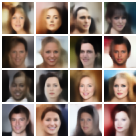}}}
\subfigure[$\mathcal{M}_G \subset \mathbb{R}^{500}$]{\includegraphics[width=0.18\textwidth]
{\FigPath{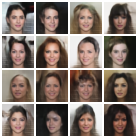}}}
\subfigure[$\mathcal{M}_G \subset \mathbb{R}^{1000}$]{\includegraphics[width=0.18\textwidth]{\FigPath{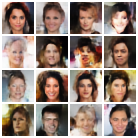}}}
\\
\subfigure[$\mathcal{M}_R \subset \mathbb{R}^{10}$]{\includegraphics[width=0.18\textwidth]{\FigPath{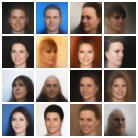}}}
\subfigure[$\mathcal{M}_R \subset \mathbb{R}^{50}$]{\includegraphics[width=0.18\textwidth]{\FigPath{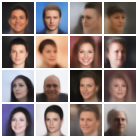}}}
\subfigure[$\mathcal{M}_R \subset \mathbb{R}^{100}$]{\includegraphics[width=0.18\textwidth]{\FigPath{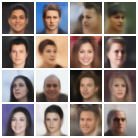}}}
\subfigure[$\mathcal{M}_R \subset \mathbb{R}^{500}$]{\includegraphics[width=0.18\textwidth]{\FigPath{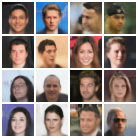}}}
\subfigure[$\mathcal{M}_R \subset \mathbb{R}^{1000}$]{\includegraphics[width=0.18\textwidth]{\FigPath{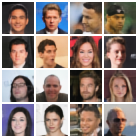}}}
\\
\subfigure[$\mathcal{M}_R \subset \mathbb{R}^{10}$]{\includegraphics[width=0.18\textwidth]{\FigPath{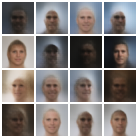}}}
\subfigure[$\mathcal{M}_R \subset \mathbb{R}^{50}$]{\includegraphics[width=0.18\textwidth]{\FigPath{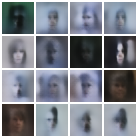}}}
\subfigure[$\mathcal{M}_R \subset \mathbb{R}^{100}$]{\includegraphics[width=0.18\textwidth]{\FigPath{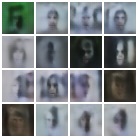}}}
\subfigure[$\mathcal{M}_R \subset \mathbb{R}^{500}$]{\includegraphics[width=0.18\textwidth]{\FigPath{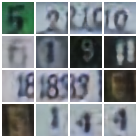}}}
\subfigure[$\mathcal{M}_R \subset \mathbb{R}^{1000}$]{\includegraphics[width=0.18\textwidth]{\FigPath{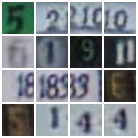}}}
\\
\subfigure[$\mathcal{M}_R \subset \mathbb{R}^{10}$]{\includegraphics[width=0.18\textwidth]{\FigPath{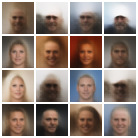}}}
\subfigure[$\mathcal{M}_R \subset \mathbb{R}^{50}$]{\includegraphics[width=0.18\textwidth]{\FigPath{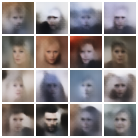}}}
\subfigure[$\mathcal{M}_R \subset \mathbb{R}^{100}$]{\includegraphics[width=0.18\textwidth]{\FigPath{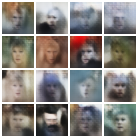}}}
\subfigure[$\mathcal{M}_R \subset \mathbb{R}^{500}$]{\includegraphics[width=0.18\textwidth]{\FigPath{learned_manifold_CelebA_500_reconstruction_cifar10.png}}}
\subfigure[$\mathcal{M}_R \subset \mathbb{R}^{1000}$]{\includegraphics[width=0.18\textwidth]{\FigPath{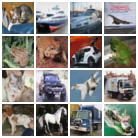}}}

\caption{
Several generated (first row) and reconstructed images (CelebA as ID data in the second row,
SVHN and CIFAR10 as OOD data in the third and forth rows, respectively)
from the proposed-D method for different manifold dimensions (10, 50, 100, 500, and 1000 from left to right) for a model trained on the CelebA dataset.
$\mathcal{M}_G \subset \mathbb{R}^{d}$ and $\mathcal{M}_R \subset \mathbb{R}^{d}$ represent image generation and image reconstruction from a manifold fall in dimension $d$, respectively.
}
\label{fig:figure_CelebA_generation_reconstruction}
\end{figure}

\subsection{Image OOD detection}
\label{sec:ood-experiment}

{
As previously outlined, our approach falls into a category of methods specifically designed to learn the underlying data distribution while maintaining robustness against OOD data, without requiring OOD samples during training.
By exclusively modeling the normal data distribution, our method inherently exhibits resilience to unseen anomalies, making it particularly effective for challenging scenarios such as:

\begin{itemize}
    \item Annotation challenges: When datasets are exceptionally large, annotating OOD instances becomes impractical due to their diversity or limited resources.
    \item Anomaly diversity: The variability of anomalies makes it infeasible to construct a comprehensive labeled dataset that accounts for all potential OOD instances.
\end{itemize}

A notable application of our method is anomaly detection in video frames \cite{LandR}, where the goal is to identify anomalies or OOD frames within a sequence. By ignoring temporal dependencies, the task can be simplified to detecting anomalies in individual frames, enabling efficient processing in many practical cases.
}

In the following sections, we first present results for OOD detection on gray-scale image datasets. Subsequently, we expand our analysis to include high-dimensional color images, demonstrating the versatility and effectiveness of our method in diverse real-world contexts.

\begin{figure}[thbp]
\centering
\subfigure[Result of MSE penalization]{\includegraphics[width=0.65\textwidth]{\FigPath{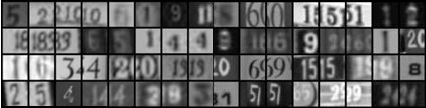}}}
\subfigure[Result of Huber penalization]{\includegraphics[width=0.32\textwidth]{\FigPath{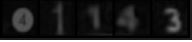}}}
\caption{
All misdetected ID $28 \times 28$ gray-scale SVHN test data for a trained model on MNIST.
}
\label{fig:grayscale_OOD}
\end{figure}

At first, the pre-defined RealNVP is employed as a backbone model for manifold learning and OOD detection of gray-scale images.
The purpose of the designed experiment is to evaluate the performance of various penalty functions in situations where distinguishing between ID and OOD data becomes challenging due to visual similarity, without considering data complexity term ($\mathbb{C}$ in \EquationRef{eq:final_loss}).
The model is trained on the MNIST dataset as ID data with an embedded manifold in $\mathbb{R}^3$. Then, it is evaluated with $28 \times 28$ gray-scale SVHN as OOD data.
A hard OOD detection threshold is determined as the maximum score obtained from the MNIST test data. Any sample with higher score than this threshold is classified as OOD data.
The reported results in \FigRef{fig:grayscale_OOD} confirm the considerable superiority of using the Huber function compared to MSE penalization in successfully identifying OOD data with high overlap.

Based on \FigRef{fig:grayscale_OOD}, the Huber function exhibits significantly fewer mistakes on the SVHN test data compared to MSE. Furthermore, the mistakes made by the Huber function are more meaningful, particularly in situations where the OOD data visually resembles the ID data (MNIST).
To provide further clarification, the Huber function nearly almost tends to make errors on images resembling data within a distribution that consists of a single digit on a dark background.
It is important to highlight that MNIST is structured in such a manner that each image contains a single digit placed on a black background.
Nevertheless, when the data deviates from this pattern, as seen in cases like SVHN (including house license plate numbers) where multiple digits may appear in a single image, the Huber function may exhibit fewer errors.
On the other hand, the MSE function exhibits a higher number of errors and even mistakenly recognizes images containing multiple digits as ID data.

{
All in all, to evaluate the effect of the regularization function ($\mathcal{C}$ in \EquationRef{eq:constrained_likelihood}), we assessed the impact of regularization functions, such as the Huber loss (\FigRef{fig:grayscale_OOD}). For instance, Huber loss outperformed MSE in detecting near-OOD cases (e.g., grayscale SVHN as OOD for MNIST) by leveraging its quadratic behavior for larger distances. Specifically, Huber penalizes small deviations linearly and larger deviations quadratically, whereas MSE applies a uniform quadratic penalty regardless of the distance. This makes Huber loss more robust. However, it struggled with SVHN samples that visually resembled MNIST, such as single digits on dark backgrounds.
}

\begin{figure}[thbp]
\centering
\subfigure[MSE penalization]{\includegraphics[width=0.44\textwidth]{\FigPath{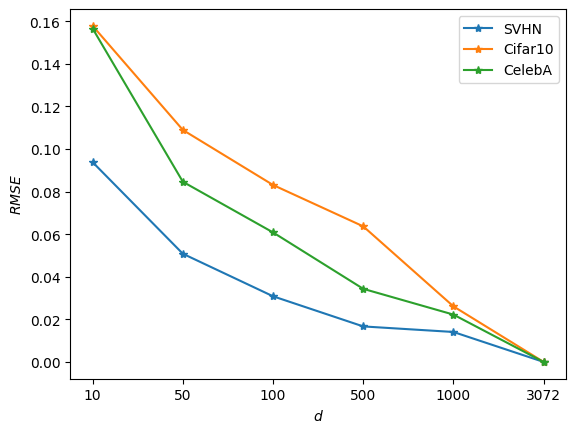}}}
\subfigure[Huber penalization]{\includegraphics[width=0.45\textwidth]{\FigPath{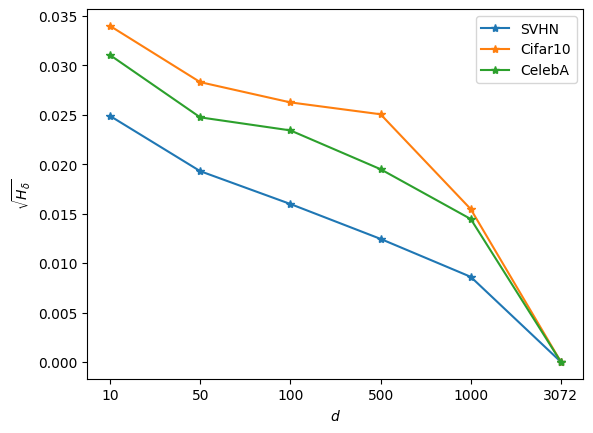}}}
\caption{
Distance from the manifold (reconstruction error) for three trained ID dataset and two penalty functions.
It is important to note that RMSE stands for the root of MSE, while $\sqrt{H_\delta}$ refers to the square root of Huber penalization, providing a more comparable measure.
}
\label{fig:reconstruction_error}
\end{figure}

In the following, we discuss the results of our proposed method for OOD detection in color images.
At first, \FigRef{fig:reconstruction_error} illustrates the reconstruction error (distance from the manifold) for two penalty functions, MSE and Huber, applied to three trained ID datasets: SVHN, Cifar10, and CelebA.
When we consider the intrinsic dimension of each ID dataset, it becomes evident that both penalty functions assign less reconstruction error to SVHN, followed by CelebA, and then Cifar10 in each dimension.
One notable observation from \FigRef{fig:reconstruction_error} is that the Huber penalty function yields a more meaningful distance from the manifold. In simpler terms, at a relatively consistent error level (approximately $0.025$), employing the Huber penalty function leads to the discovery of lower-dimensional manifolds, whereas achieving a similar error level with MSE would require higher manifold dimensions.

We compare our proposed method for color image OOD detection with other state-of-the-art methods \citep{kingma2018glow,ren2019likelihood, Serrà2020Input}.
The important reason for selecting the methods to compare in the following experiments is their alignment with the assumptions of our proposed method. This entails not using auxiliary OOD training data, not customizing the model structure for this specific task, and maintaining a likelihood-based generative model approach while still achieving state-of-the-art results. To ensure a fair comparison, we re-implement all these methods using the same backbone model (the architecture detailed in Section \SectionRef{sec:architecture}).

For the dimension-preserving experiment ($d=D$), we evaluate the following methods: the baseline method \citep{kingma2018glow} (named Glow, using the architecture detailed in Section \SectionRef{sec:architecture}), the likelihood ratio method \citep{ren2019likelihood} (named ratio, employing the same architecture as the reported Glow network in Section \SectionRef{sec:architecture}), and the input complexity method \citep{Serrà2020Input} (named IC, with the architecture matching that of the reported Glow network in Section \SectionRef{sec:architecture}).
We also integrate the proposed idea into another state-of-the-art method, achieving valuable results in OOD detection. These combinations include:
the proposed method on Glow architecture (labeled as $\text{P}$ for \EquationRef{eq:loss});
the proposed method with consideration of input complexity (labeled as $\text{P}+\text{IC}$ for \EquationRef{eq:final_loss}).
To summarize our approach to integrating the proposed method with IC, it is important to note that, as explained in detail in Section \SectionRef{sec:related_work} and \EquationRef{eq:IC_loss}, this method is used as test-time score. We incorporate only the compressor loss from the input complexity method into our final score as explained in \EquationRef{eq:final_loss}.

{
Before presenting the experimental results, it is important to note that the complexity of the architecture, which affects the processing time of a method, is crucial for evaluating its feasibility.
Our proposed framework falls under the category of test-time evaluation methods, similar to state-of-the-art techniques such as ratio and IC, and offers significant computational advantages:
\begin{itemize}
    \item Ratio: This method requires training an auxiliary background model alongside the backbone model, which introduces substantial training overhead. During inference, both the backbone and auxiliary models must be evaluated, resulting in considerable inference overhead.
    
    \item IC: This method avoids training an auxiliary model but relies on a pre-trained compressor. During inference, data must pass through the compressor, increasing computational demands.

    \item Proposed method (P): The core of our approach lies in modifying the training process to encode a manifold representation within part of the latent space. This enables distance measurement directly in this space, which can be computed using criteria such as MSE or Huber functions.
    Our method eliminates the need for direct auxiliary models or compressors; however, it requires the backward pass of the NF, which is time-consuming. Fortunately, it uses the inverse of the NF (without requiring additional training), keeping the overhead minimal, though not negligible.
    More details about the inference time will be discussed at the end of the section.
    
    \item Proposed method integrated with IC (P+IC): When combined with IC, our method uses the IC compressor during inference. However, similar to method P, the overhead from the backward pass adds to this, making it the most time-consuming method in this scenario.
\end{itemize}
}

The effect of our proposed method on color image OOD detection for NFs is evaluated in terms of AUROC (Area Under the Receiver Operating Characteristic Curve) criterion.
AUROC computes the area under the ROC curve by using un-thresholded normalized predictions to the range of $[0, 1]$
on the data along with binary true labels. In our configuration, we assign a true label of 0 for ID data and a true label of 1 for OOD data.
A higher AUROC value indicates better performance.

Finally, we report the comprehensive results of the proposed method and selected competitors for two different distance levels from the manifold in \TableRef{tab:auroc1} and \TableRef{tab:auroc2}. \TableRef{tab:auroc1} corresponds to trained models with RMSE=0.1 and $\sqrt{H_\delta}=0.025$, while \TableRef{tab:auroc2} is associated with models with RMSE=0.02 and $\sqrt{H_\delta}=0.005$.
As indicated in \FigRef{fig:reconstruction_error}, the dimension of the selected manifold varies for each dataset. Therefore, the models reported in \TableRef{tab:auroc1} and \TableRef{tab:auroc2} are determined based on the dimension at which the models achieved the specified distance level.
In \TableRef{tab:auroc1}, the manifold dimensions associated with the trained model on SVHN, CelebA, and Cifar10 are 10, 50, and 100, respectively. For \TableRef{tab:auroc2}, these values have been updated to 500, 1000, and 1000, correspondingly.

Both \TableRef{tab:auroc1} and \TableRef{tab:auroc2} share the same structure, with the first segment of each table corresponding to the MSE penalization function, and the second part relating to the Huber function.
The first column of tables provides the names of the OOD data used during testing. Subsequent columns represent different trained model of ID datasets (SVHN, Cifar10, and CelebA) and maintain consistent partitioning, including:
the likelihood ratio method \citep{ren2019likelihood} with an architecture matching the reported Glow network in \SectionRef{sec:architecture};
results from the baseline method \citep{kingma2018glow} with the architecture reported in \SectionRef{sec:architecture};
the input complexity method \citep{Serrà2020Input} with an architecture identical to the reported Glow network in \SectionRef{sec:architecture};
the performance of our proposed method;
the performance of our proposed method combined with the IC method, denoted as $\text{P}+\text{IC}$.

\begin{landscape}

\begin{table}[h]
\setlength{\tabcolsep}{4pt}
\centering
\caption{AUROC score for likelihood ratio method (named ratio) \citep{ren2019likelihood}, Glow \citep{kingma2018glow}, input complexity method (named IC) \citep{Serrà2020Input} with a lossless PNG compressor, and the best performance of the proposed method (named P and $\text{P}+\text{IC}$) for MSE and Huber penalization.
The top section of the table displays the outcomes of the MSE penalty, while the bottom section contains the results for the Huber penalty.
The models are obtained based on the distance level from the manifold for RMSE=0.1 and $\sqrt{H_\delta}=0.025$.
Given the specified distance from the manifold, the manifold dimensions for the trained model on SVHN, CelebA, and Cifar10 are 10, 50, and 100, respectively.
}
\label{tab:auroc1}
\begin{tabularx}{\linewidth}{l*{15}{>{\centering\arraybackslash}p{0.9cm}}}
	\cmidrule(lr){1-16}
	Data & \multicolumn{5}{c}{SVHN} & \multicolumn{5}{c}{Cifar10} & \multicolumn{5}{c}{CelebA} \\
	\cmidrule(lr){2-6} \cmidrule(lr){7-11} \cmidrule(lr){12-16}
	& ratio & Glow & IC & P & P+IC & ratio & Glow & IC & P & P+IC & ratio & Glow & IC & P & P+IC \\
	\cmidrule(lr){1-16}
	
    MNIST
    & $0.588$ & $0.662$ & $1.0$ & $0.958$ & $1.0$
    &  $0.004$ & $0.002$ & $1.0$ & $0.331$ & $1.0$
    & $0.001$ & $0.0$ & $1.0$ & $0.876$ & $1.0$ \\

    FMNIST
    & $0.998$ & $0.863$ & $1.0$ & $0.99$ & $1.0$
    & $0.014$ & $0.016$ & $0.999$ & $0.455$ & $0.999$
    & $0.004$ & $0.013$ & $1.0$ & $0.791$ & $1.0$ \\

    SVHN
    & - & - & - & - & -
    & $0.004$ & $0.057$ & $0.86$ & $0.056$ & $0.831$
    & $0.624$ & $0.7$ & $0.655$ & $0.194$ & $0.917$ \\

    Cifar10
    & $1.0$ & $0.99$ & $0.408$ & $0.989$ & $0.938$
    & - & - & - & - & -
    & $0.619$ & $0.698$ & $0.871$ & $0.814$ & $0.841$ \\

    LSUN
    & $1.0$ & $0.996$ & $0.647$ & $0.994$ & $0.981$
    & $0.544$ & $0.541$ & $0.779$ & $0.506$ & $0.785$
    & $0.0$ & $0.061$ & $0.935$ & $0.829$ & $0.953$ \\

    CelebA
    & $1.0$ & $0.998$ & $0.423$ & $0.997$ & $0.978$
    & $0.558$ & $0.486$ & $0.558$ & $0.498$ & $0.558$
    & - & - & - & - & - \\

    \rowcolor{Gray}
    Average
    & $0.917$ & $0.902$ & $0.696$ & $0.986$ & $0.979$
    & $0.225$ & $0.220$ & \textbf{$0.839$} & $0.369$ & $0.835$
    & $0.250$ & $0.294$ & $0.892$ & $0.701$ & \textbf{$0.942$}
    \\
	
	\cmidrule(lr){1-16}
	
	  MNIST
	  & $0.588$ & $0.662$ & $1.0$ & $0.928$ & $1.0$
	  &  $0.004$ & $0.002$ & $1.0$ & $0.477$ & $1.0$
	  & $0.001$ & $0.0$ & $1.0$ & $0.898$ & $1.0$ \\
	
	  FMNIST
	  & $0.998$ & $0.863$ & $1.0$ & $0.955$ & $1.0$
	  & $0.014$ & $0.016$ & $0.999$ & $0.543$ & $0.999$
	  & $0.004$ & $0.013$ & $1.0$ & $0.836$ & $1.0$ \\
	
	  SVHN
	  & - & - & - & - & -
	  & $0.004$ & $0.057$ & $0.86$ & $0.111$ & $0.861$
	  & $0.624$ & $0.7$ & $0.655$ & $0.283$ & $0.919$ \\
	
	  Cifar10
	  & $1.0$ & $0.99$ & $0.408$ & $0.98$ & $0.947$
	  & - & - & - & - & -
	  & $0.619$ & $0.698$ & $0.871$ & $0.831$ & $0.89$ \\
	
	  LSUN
	  & $1.0$ & $0.996$ & $0.647$ & $0.988$ & $0.984$
	  & $0.544$ & $0.541$ & $0.779$ & $0.539$ & $0.775$
	  & $0.0$ & $0.061$ & $0.935$ & $0.855$ & $0.966$ \\
	
	  CelebA
	  & $1.0$ & $0.998$ & $0.423$ & $0.989$ & $0.975$
	  & $0.558$ & $0.486$ & $0.558$ & $0.581$ & $0.576$
	  & - & - & - & - & - \\
	
	  \rowcolor{Gray}
	  Average
	  & $0.917$ & $0.902$ & $0.696$ & $0.97$ & $0.981$
	  & $0.225$ & $0.220$ & $0.839$ & $0.45$ & \textbf{$0.842$}
	  & $0.250$ & $0.294$ & $0.892$ & $0.741$ & \textbf{$0.956$} \\
	
\end{tabularx}
\end{table}

\begin{table}[h]
\setlength{\tabcolsep}{4pt}
\centering
\caption{AUROC score for likelihood ratio method (named ratio) \citep{ren2019likelihood}, Glow \citep{kingma2018glow}, input complexity method (named IC) \citep{Serrà2020Input} with a lossless PNG compressor, and the best performance of the proposed method (named P and $\text{P}+\text{IC}$) for MSE and Huber penalization.
The top section of the table displays the outcomes of the MSE penalty, while the bottom section contains the results for the Huber penalty.
The models are obtained based on the distance level from the manifold for RMSE=0.02 and $\sqrt{H_\delta}=0.005$.
Given the specified distance from the manifold, the manifold dimensions for the trained model on SVHN, CelebA, and Cifar10 are 500, 1000, and 1000, respectively.
}
\label{tab:auroc2}
\begin{tabularx}{\linewidth}{l*{15}{>{\centering\arraybackslash}p{0.9cm}}}
	\cmidrule(lr){1-16}
	Data & \multicolumn{5}{c}{SVHN} & \multicolumn{5}{c}{Cifar10} & \multicolumn{5}{c}{CelebA} \\
	\cmidrule(lr){2-6} \cmidrule(lr){7-11} \cmidrule(lr){12-16}
	& ratio & Glow & IC & P & P+IC & ratio & Glow & IC & P & P+IC & ratio & Glow & IC & P & P+IC \\
	\cmidrule(lr){1-16}
	
    MNIST
    & $0.588$ & $0.662$ & $1.0$ & $0.999$ & $1.0$
    &  $0.004$ & $0.002$ & $1.0$ & $0.884$ & $0.999$
    & $0.001$ & $0.0$ & $1.0$ & $0.996$ & $1.0$ \\
    
    FMNIST
    & $0.998$ & $0.863$ & $1.0$ & $0.997$ & $0.999$
    & $0.014$ & $0.016$ & $0.999$ & $0.758$ & $0.998$
    & $0.004$ & $0.013$ & $1.0$ & $0.957$ & $1.0$ \\
    
    SVHN
    & - & - & - & - & -
    & $0.004$ & $0.057$ & $0.86$ & $0.055$ & $0.702$
    & $0.624$ & $0.7$ & $0.655$ & $0.072$ & $0.933$ \\
    
    Cifar10
    & $1.0$ & $0.99$ & $0.408$ & $0.996$ & $0.997$
    & - & - & - & - & -
    & $0.619$ & $0.698$ & $0.871$ & $0.665$ & $0.679$ \\
    
    LSUN
    & $1.0$ & $0.996$ & $0.647$ & $0.997$ & $0.998$
    & $0.544$ & $0.541$ & $0.779$ & $0.566$ & $0.759$
    & $0.0$ & $0.061$ & $0.935$ & $0.683$ & $0.879$ \\
    
    CelebA
    & $1.0$ & $0.998$ & $0.423$ & $0.998$ & $0.998$
    & $0.558$ & $0.486$ & $0.558$ & $0.531$ & $0.544$
    & - & - & - & - & - \\
    
    \rowcolor{Gray}
    Average
    & $0.917$ & $0.902$ & $0.696$ & $0.997$ &  \textbf{$0.998$}
    & $0.225$ & $0.220$ &  \textbf{$0.839$} & $0.559$ & $0.8$
    & $0.250$ & $0.294$ & $0.892$ & $0.675$ &  \textbf{$0.898$}
    \\
	
	\cmidrule(lr){1-16}
	
	 MNIST
	 & $0.588$ & $0.662$ & $1.0$ & $0.986$ & $1.0$
	 &  $0.004$ & $0.002$ & $1.0$ & $0.916$ & $1.0$
	 & $0.001$ & $0.0$ & $1.0$ & $0.992$ & $1.0$ \\
	 
      FMNIST
      & $0.998$ & $0.863$ & $1.0$ & $0.99$ & $1.0$
	 & $0.014$ & $0.016$ & $0.999$ & $0.819$ & $0.999$
	 & $0.004$ & $0.013$ & $1.0$ & $0.933$ & $1.0$ \\
	 
	 SVHN
	 & - & - & - & - & -
	 & $0.004$ & $0.057$ & $0.86$ & $0.068$ & $0.849$
	 & $0.624$ & $0.7$ & $0.655$ & $0.125$ & $0.919$ \\
	 
	 Cifar10
	 & $1.0$ & $0.99$ & $0.408$ & $0.993$ & $0.996$
	 & - & - & - & - & -
	 & $0.619$ & $0.698$ & $0.871$ & $0.976$ & $0.701$ \\
	 
      LSUN
	 & $1.0$ & $0.996$ & $0.647$ & $0.996$ & $0.999$
	 & $0.544$ & $0.541$ & $0.779$ & $0.552$ & $0.781$
	 & $0.0$ & $0.061$ & $0.935$ & $0.681$ & $0.888$ \\
	 
	 CelebA
	 & $1.0$ & $0.998$ & $0.423$ & $0.998$ & $0.998$
	 & $0.558$ & $0.486$ & $0.558$ & $0.563$ & $0.559$
	 & - & - & - & - & - \\
	 
	 \rowcolor{Gray}
	 Average
	 & $0.917$ & $0.902$ & $0.696$ & $0.993$ & \textbf{$0.999$}
	 & $0.225$ & $0.220$ & \textbf{$0.839$} & $0.584$ & $0.837$
	 & $0.250$ & $0.294$ & $0.892$ & $0.741$ & \textbf{$0.902$} \\
	
\end{tabularx}
\end{table}

\end{landscape}

Both reported tables show significant performance, which we will analyze. It is important to note that the three columns (ratio, Glow, IC) are independent of the penalization function, as they are dimension-preserving methods. Consequently, their performance remains consistent across the two different penalization functions for each ID data. As mentioned earlier, all three methods share the same underlying structure to ensure comparability.
The key observation is that the ratio method performs well on data with higher visual complexity, thanks to its background model, which distinguishes between the background and semantic content in addition to the density estimation model.
Conversely, the IC method excels on simple structured data when considering data complexity.
Moreover, Glow as an NF method that does not learn the manifold, may not consistently offer interpretability for different data types. This inconsistency arises from likelihood ambiguity.
For our proposed method, we employed two distinct approaches. First, we applied manifold learning to Glow (referred to as P in the tables), resulting in a significant improvement in likelihood for OOD detection. As evident from the results, on average, this approach outperforms the baseline methods. Additionally, to address data complexity (similar to IC), we incorporated data complexity into the proposed method (named P+IC), ultimately yielding the best results with both advantages.

All in all, it is important to note that using Huber penalty function generally leads to an average improvement {in the AUROC metric} compared to using MSE.
Moreover, the results are robust and do not change significantly for different distance levels from the manifold.
{ Briefly, as shown in \TableRef{tab:auroc1} and \TableRef{tab:auroc2}, for well-structured manifolds, the proposed approach consistently outperformed existing methods. However, for complex datasets like Cifar10, the proposed method which is capable of modeling a single-chart manifold proved insufficient, highlighting the possible need for multi-manifold modeling. To address the current limitations, we incorporated data complexity modeling inspired by \citep{Serrà2020Input}, establishing a foundation for a multi-manifold solution in future studies.}

{
The evaluation results so far have primarily focused on AUROC, which is a widely accepted metric for OOD detection. To provide a more comprehensive assessment, the model's performance is also analyzed using the corresponding Receiver Operating Characteristic (ROC) curves to better visualize false positive rates versus true positive rates. These curves are presented in the \FigRef{fig:metrics} and support our insights.

\begin{landscape}

\begin{figure}[t]
     \centering     
     \subfigure[SVHN from \TableRef{tab:auroc1}]
     {
     \includegraphics[width=0.45\textwidth]         {\FigPath{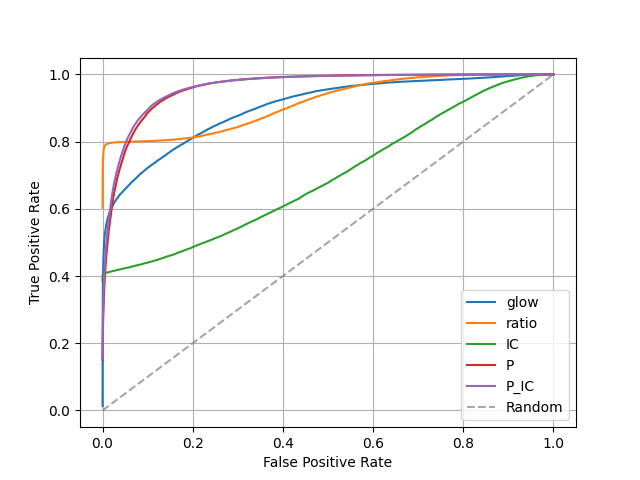}}
     }
     ~
     \subfigure[Cifar10 from \TableRef{tab:auroc1}]
     {
     \includegraphics[width=0.45\textwidth]         {\FigPath{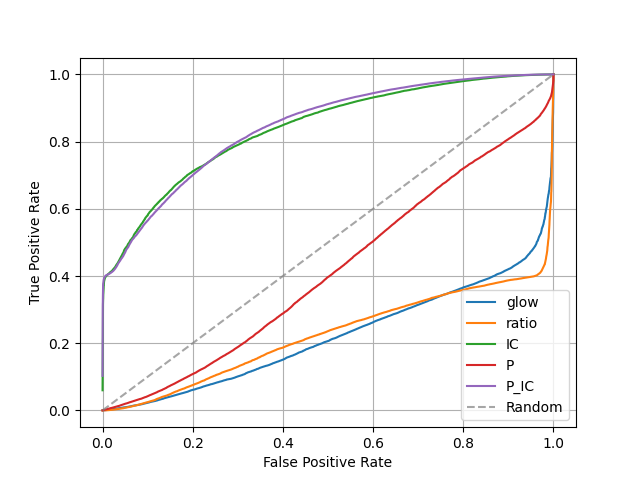}}
     }
     ~
     \subfigure[CelebA from \TableRef{tab:auroc1}]
     {
     \includegraphics[width=0.45\textwidth]         {\FigPath{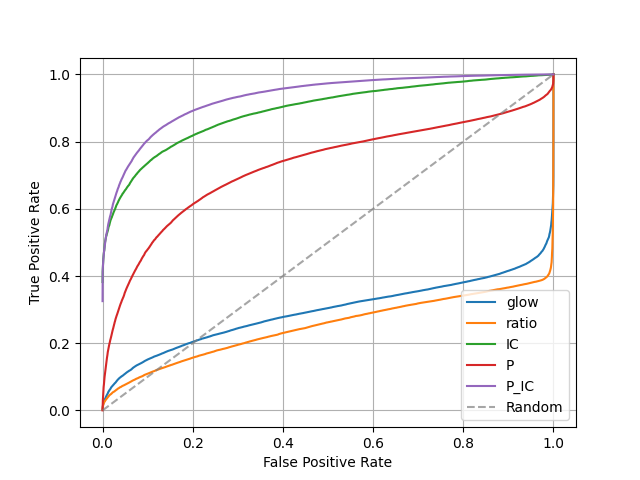}}
     }
     \\
     \subfigure[SVHN from \TableRef{tab:auroc2}]
     {
     \includegraphics[width=0.45\textwidth]         {\FigPath{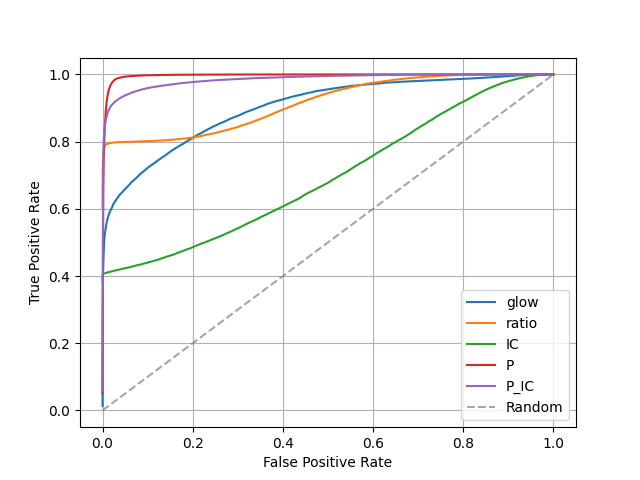}}
     }
     ~
     \subfigure[Cifar10 from \TableRef{tab:auroc2}]
     {
     \includegraphics[width=0.45\textwidth]         {\FigPath{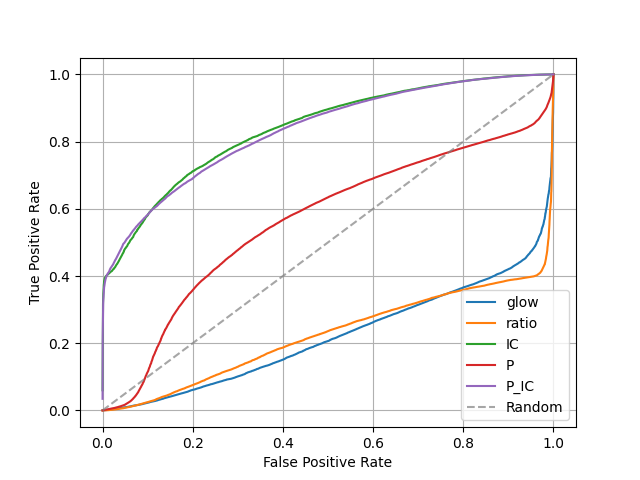}}
     }
     ~
     \subfigure[CelebA from \TableRef{tab:auroc2}]
     {
     \includegraphics[width=0.45\textwidth]         {\FigPath{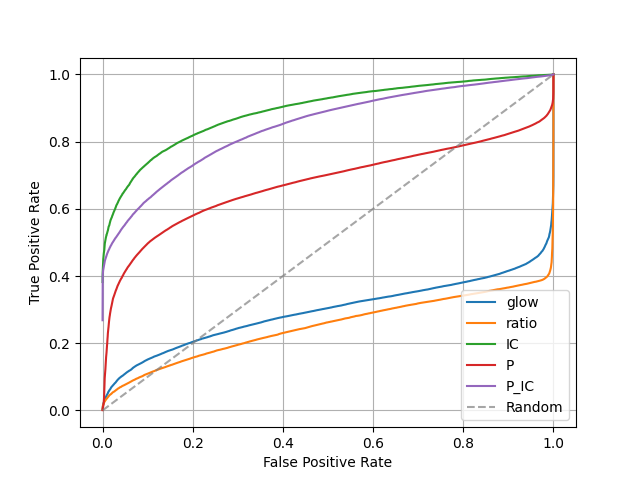}}
     }     
    \caption{
    The ROC plots corresponding to \TableRef{tab:auroc1} and \TableRef{tab:auroc2}. Considering that we have three ID datasets, each paired with five OOD datasets, and the proposed models are trained with two different penalty functions, the mean ROC values are plotted for each column to create concise and meaningful graphs. Consequently, each part of the tables is visualized as three independent plots. The first row of this figure corresponds to the results from \TableRef{tab:auroc1}, while the second row corresponds to \TableRef{tab:auroc2}.
    }
    \label{fig:metrics}
\end{figure}

\end{landscape}

}

{
Beyond presenting the results above, we deemed it crucial to include an analysis of runtime performance in this paper, comparing our methods with similar approaches \citep{ren2019likelihood, Serrà2020Input}.
The ratio, IC, and our proposed methods (P, P+IC) are all test-time OOD detection techniques built on a shared backbone: the Glow model \citep{kingma2018glow}. However, as summarized in the \TableRef{tab:times}, some methods incorporate an auxiliary model, which may or may not require training (refer to columns \textit{Auxiliary model} and \textit{Training required?}).
Details of the used system for evaluation are provided in the \SectionRef{sec:training}.

While the computational framework (Glow backbone) remains consistent across all methods, the primary variation in training time derives from the auxiliary model. Specifically, training time is influenced by whether the auxiliary model requires training and the extent of that training.
Regarding the two proposed methods (P and P+IC), although we do not employ a direct auxiliary model, the calculation of the reconstruction loss requires performing the backward pass of the used NF.

In terms of inference, we assessed the average runtime during inference for each method in milliseconds (ms), based on five independent runs, as shown in \TableRef{tab:times}. These evaluations were performed on the entire test dataset using models trained on three different ID datasets: SVHN, Cifar10, and CelebA.
The results show that IC achieves the shortest runtime, as it relies only on the shared backbone and a pre-trained auxiliary model (standard image compressor models). In contrast, P+IC has the longest runtime among the methods, as it requires both the backward pass and the use of the compressor integrated with IC. Another proposed method (referred to as P) demonstrates a shorter runtime compared to P+IC, as it does not involve a compressor. The ratio method exhibits a moderate average runtime, likely due to variations in the size of the auxiliary model used.
In conclusion, it is worth mentioning that although our best proposed method (P+IC) requires more time compared to the IC method, as previously observed, its performance is comparable and, in some cases, superior in identifying OOD data. Moreover, it possesses the capability to learn the data manifold effectively.

\begin{table}[h]

\setlength{\tabcolsep}{4pt}
\centering
\caption{
Comparison of runtime performance for test-time OOD detection methods (ratio \citep{ren2019likelihood}, IC \citep{Serrà2020Input}, P, and P+IC) using the shared Glow \citep{kingma2018glow} backbone. The table also shows whether each method incorporates an auxiliary model, and the auxiliary method requires training.
Note that the inference times represent the average of five independent runs, rounded to two decimal places.}
\label{tab:times}
\begin{tabular}{|l|c|c|c|}
\hline
\rowcolor{Gray} \textbf{Method} & \textbf{Auxiliary model} & \textbf{Training required?} & \textbf{Inference Time (ms)} \\ \hline
ratio        & \cmark & \cmark & 2.30 \\ 
IC           & \cmark & \xmark & 1.44 \\ 
P   & \xmark & \xmark & 2.20 \\ 
P+IC    & \cmark & \xmark & 2.55 \\
\hline
\end{tabular}
\end{table}
}

{
Scalability is an important aspect of the proposed OOD detection method, especially as the dimensionality and the complexity of data  increase. Likelihood-based approaches often face challenges in high-dimensional spaces, where they tend to assign high probabilities to irrelevant regions. The dimensionality of the data used in our experiments was 3072 which is considered high for fitting probability densities, and as we observed in the experiments, the pure likelihood-based method exhibited a poor performance.
Our method tackles this issue by leveraging the manifold hypothesis. It focuses on lower-dimensional manifolds where real-world data resides, avoiding inefficiencies in modeling the entire observation space. This ensures scalability and can also be extended to multi-manifold settings, allowing effective modeling of data embedded in multiple distinct manifolds.
Unlike other methods like  ratio and IC, which require auxiliary models to tackle scalability problem, our method possesses intrinsic properties that ensure scalability without relying on external tools.}

\section{Conclusion}
\label{sec:conclusion}
NFs cannot directly provide an estimation of the likelihood on the data manifold. To address this, we adopted an algorithmic approach to jointly estimate the likelihood and data manifold, and used it for OOD detection.
Our proposed method is based on the intuition that relying only on likelihood value on the manifold or distance from the manifold is insufficient for OOD detection.
Accordingly, we found that using these two indicators, namely likelihood and distance from the manifold, jointly significantly improves OOD detection.
In this paper, a per-pixel (an element-wise) penalization function (MSE, Huber) was used for manifold learning.
MSE is the usual choice for manifold learning, while the Huber function was employed here due to its transition from linear to quadratic mode.
It penalizes off-manifold parts linearly, and the model focuses on learning the on-manifold sub-space well.
Moreover, we employed a prior method to obtain a likelihood estimator based on data complexity, that enabled us to refine the model's likelihood. This enhancement of our proposed method yielded significant results.
Through extensive experiments, we showed the strength of the proposed method in finding data manifold and estimating its likelihood, besides showing excellent performance for image OOD detection.
Our approach, which applies manifold learning to NFs, significantly improved OOD detection and outperformed baseline methods. Further enhancement with data complexity led to the best results. The use of the Huber penalty function generally resulted in improved performance compared to MSE.
The proposed method can also generate samples similar to the latest state-of-the-art manifold learning using NFs in terms of visual quality.

As a future work, per-object (a group of pixels) and per-sample (all pixels) penalization can be an interesting work to follow in a future work.
Also, it should be noted that a primary assumption of the current study is that data lie on a manifold with one mapping (or 'chart' in manifold literature).
In future work, we aim to extend a multi-mapping (or 'multi-chart' in manifold literature) version of the current method with possibly improved ability of OOD detection.
As another potential future research directions, it is worthwhile to investigate whether the ratio of on-manifold to off-manifold points remains consistent across all manifold dimensions.
In the context of our current study, we posited that in lower dimensions, it is essential to place greater emphasis on modeling on-manifolds. In other words, as the distance from the data manifold increases, the influence of the linear criteria in the Huber loss (Laplace distribution) grows to mitigate significant penalties.
This approach was facilitated by selecting an appropriate $\delta$, especially in low dimensions.
Nevertheless, there exists potential for exploring the effects of equal proportion by examining the contributions of on-manifold and off-manifold points.
As another future work suggestion, addressing the homogenization of terms in the proposed score is crucial. Achieving this homogeneity involves determining a coefficient based on the variance of the manifold distance distribution, aligning it with a likelihood-based concept. While our focus has been on the inference phase, this can also be integrated during training to ensure consistency in the incorporation of these terms. We anticipate that implementing this change will enhance the training process and boost OOD detection performance.
{
As a final suggestion for future work to reduce the computation time of the proposed method. In our method, calculating the distance from the manifold involves a backward pass, which nearly doubles the computational complexity. To address this, we propose exploring the use of a smaller auxiliary model that could replace the backward pass, reducing computation time while potentially enhancing OOD detection performance.
}

\begin{appendices}
\section{Huber density function}
\label{sec:appendix}

In this appendix, we want to incorporate a scale in the definition of Huber density function, and explain how this scale parameter can be estimated using maximum likelihood. In the literature, the Huber density is defined as
\begin{equation}
	P_\delta(e) = C_{\delta} \exp \big( -H_\delta(e) \big),
 \label{eq:general_huber_density}
\end{equation}
where $H_\delta(e)$ is the Huber error function, that is
\begin{equation}
\label{eq:hubererror}
    H_\delta(e) = 
     \begin{cases}
      \frac12 e^2, & \text{if } |e| < \delta,
      \\
      \delta(|e| - \frac12 \delta), & \text{otherwise},  \\ 
     \end{cases}
\end{equation}
and the normalization constant ($C_\delta$) is calculated using
\begin{equation}
\label{eq:huber_density_criteria}
	C_{\delta} = 
	\Big(
	\frac{2}{\delta}
	\exp(\frac{-\delta^2}{2})
	+
	\sqrt{2\pi}
	\big(
	2\Phi(\delta) - 1
	\big)
	\Big)^{-1},
\end{equation}
where $\Phi$ is the error function.

For incorporating scale, we transform the variable using $e^\prime = g(e) = ke$, for positive $k$. The density in the transformed space is
\begin{equation}
        \label{eq:change_of_variable_formula}
        P_{e^\prime}(e^\prime; \delta, k)
        =
        P_{e}\biggl(\frac{e^\prime}{k}; \delta\biggr) \frac{1}{k},
\end{equation}
and therefore we obtain
\begin{equation}
\label{eq:huber_density}
P_{e^\prime}(e^\prime; \delta^\prime,k) = C_{\delta^\prime, k} \exp \bigg( \frac{-H_{\delta^\prime}(e^\prime)}{k^2} \bigg),
\end{equation}
where $\delta^\prime = k \delta$ and
\begin{equation}
	C_{\delta^\prime, k} = 
	\Big(
	\frac{2k^2}{\delta^\prime}
	\exp(\frac{-\delta^{\prime^2}}{2k^2})
	+
	\sqrt{2\pi}k
	\big(
	2\Phi(\frac{\delta^\prime}{k}) - 1
	\big)
	\Big)^{-1}.
\end{equation}

In the Huber density, the value of $k^2$ is proportional to the density's variance and our goal is to estimate this parameter. Given a set of errors and a specific $\delta^\prime$, we are looking for the optimal $k$ that minimizes the NLL, that is
\begin{equation}
    k^{*} = \arg \underset{k}{\min} \bigg(- N \log (C_{\delta^\prime,k})
	+ \frac{\sum_{n=1}^{n=N} H_{\delta^\prime}(e^\prime_n)}{k^2}\bigg).
    \label{eq:nll_huber}
\end{equation}

A closed-form solution is not accessible for \EquationRef{eq:nll_huber}, but it is a simple univariate optimization. As mentioned in the paper, it is tackled through numerical optimization techniques, specifically employing the Newton method.

\end{appendices}

%
%
%
%
%
%

\bibliography{sn-bibliography}


\begin{thebibliography}{47}
\ifx \bisbn   \undefined \def \bisbn  #1{ISBN #1}\fi
\ifx \binits  \undefined \def \binits#1{#1}\fi
\ifx \bauthor  \undefined \def \bauthor#1{#1}\fi
\ifx \batitle  \undefined \def \batitle#1{#1}\fi
\ifx \bjtitle  \undefined \def \bjtitle#1{#1}\fi
\ifx \bvolume  \undefined \def \bvolume#1{\textbf{#1}}\fi
\ifx \byear  \undefined \def \byear#1{#1}\fi
\ifx \bissue  \undefined \def \bissue#1{#1}\fi
\ifx \bfpage  \undefined \def \bfpage#1{#1}\fi
\ifx \blpage  \undefined \def \blpage #1{#1}\fi
\ifx \burl  \undefined \def \burl#1{\textsf{#1}}\fi
\ifx \doiurl  \undefined \def \doiurl#1{\url{https://doi.org/#1}}\fi
\ifx \betal  \undefined \def \betal{\textit{et al.}}\fi
\ifx \binstitute  \undefined \def \binstitute#1{#1}\fi
\ifx \binstitutionaled  \undefined \def \binstitutionaled#1{#1}\fi
\ifx \bctitle  \undefined \def \bctitle#1{#1}\fi
\ifx \beditor  \undefined \def \beditor#1{#1}\fi
\ifx \bpublisher  \undefined \def \bpublisher#1{#1}\fi
\ifx \bbtitle  \undefined \def \bbtitle#1{#1}\fi
\ifx \bedition  \undefined \def \bedition#1{#1}\fi
\ifx \bseriesno  \undefined \def \bseriesno#1{#1}\fi
\ifx \blocation  \undefined \def \blocation#1{#1}\fi
\ifx \bsertitle  \undefined \def \bsertitle#1{#1}\fi
\ifx \bsnm \undefined \def \bsnm#1{#1}\fi
\ifx \bsuffix \undefined \def \bsuffix#1{#1}\fi
\ifx \bparticle \undefined \def \bparticle#1{#1}\fi
\ifx \barticle \undefined \def \barticle#1{#1}\fi
\bibcommenthead
\ifx \bconfdate \undefined \def \bconfdate #1{#1}\fi
\ifx \botherref \undefined \def \botherref #1{#1}\fi
\ifx \url \undefined \def \url#1{\textsf{#1}}\fi
\ifx \bchapter \undefined \def \bchapter#1{#1}\fi
\ifx \bbook \undefined \def \bbook#1{#1}\fi
\ifx \bcomment \undefined \def \bcomment#1{#1}\fi
\ifx \oauthor \undefined \def \oauthor#1{#1}\fi
\ifx \citeauthoryear \undefined \def \citeauthoryear#1{#1}\fi
\ifx \endbibitem  \undefined \def \endbibitem {}\fi
\ifx \bconflocation  \undefined \def \bconflocation#1{#1}\fi
\ifx \arxivurl  \undefined \def \arxivurl#1{\textsf{#1}}\fi
\csname PreBibitemsHook\endcsname

\bibitem[\protect\citeauthoryear{Yang et~al.}{2024}]{yang2021generalized}
\begin{barticle}
\bauthor{\bsnm{Yang}, \binits{J.}},
\bauthor{\bsnm{Zhou}, \binits{K.}},
\bauthor{\bsnm{Li}, \binits{Y.}},
\bauthor{\bsnm{Liu}, \binits{Z.}}:
\batitle{Generalized out-of-distribution detection: {A} survey}.
\bjtitle{International Journal of Computer Vision}
\bvolume{132}(\bissue{12}),
\bfpage{5635}--\blpage{5662}
(\byear{2024})
\end{barticle}
\endbibitem

\bibitem[\protect\citeauthoryear{Nalisnick et~al.}{2019}]{nalisnick2018deep}
\begin{bchapter}
\bauthor{\bsnm{Nalisnick}, \binits{E.T.}},
\bauthor{\bsnm{Matsukawa}, \binits{A.}},
\bauthor{\bsnm{Teh}, \binits{Y.W.}},
\bauthor{\bsnm{G{\"{o}}r{\"{u}}r}, \binits{D.}},
\bauthor{\bsnm{Lakshminarayanan}, \binits{B.}}:
\bctitle{Do deep generative models know what they don't know?}
In: \bbtitle{International Conference on Learning Representations}.
\bpublisher{OpenReview.net},
\blocation{Los Angeles}
(\byear{2019})
\end{bchapter}
\endbibitem

\bibitem[\protect\citeauthoryear{Kirichenko
  et~al.}{2020}]{kirichenko2020normalizing}
\begin{bchapter}
\bauthor{\bsnm{Kirichenko}, \binits{P.}},
\bauthor{\bsnm{Izmailov}, \binits{P.}},
\bauthor{\bsnm{Wilson}, \binits{A.G.}}:
\bctitle{Why normalizing flows fail to detect out-of-distribution data}.
In: \bbtitle{Advances in Neural Information Processing Systems},
vol. \bseriesno{33},
pp. \bfpage{20578}--\blpage{20589}.
\bpublisher{Curran Associates, Inc.},
\blocation{Virtual}
(\byear{2020})
\end{bchapter}
\endbibitem

\bibitem[\protect\citeauthoryear{Theis et~al.}{2016}]{theis2015note}
\begin{bchapter}
\bauthor{\bsnm{Theis}, \binits{L.}},
\bauthor{\bsnm{Oord}, \binits{A.}},
\bauthor{\bsnm{Bethge}, \binits{M.}}:
\bctitle{A note on the evaluation of generative models}.
In: \bbtitle{International Conference on Learning Representations},
\bconflocation{San Juan}
(\byear{2016})
\end{bchapter}
\endbibitem

\bibitem[\protect\citeauthoryear{Ren et~al.}{2019}]{ren2019likelihood}
\begin{bchapter}
\bauthor{\bsnm{Ren}, \binits{J.}},
\bauthor{\bsnm{Liu}, \binits{P.J.}},
\bauthor{\bsnm{Fertig}, \binits{E.}},
\bauthor{\bsnm{Snoek}, \binits{J.}},
\bauthor{\bsnm{Poplin}, \binits{R.}},
\bauthor{\bsnm{Depristo}, \binits{M.}},
\bauthor{\bsnm{Dillon}, \binits{J.}},
\bauthor{\bsnm{Lakshminarayanan}, \binits{B.}}:
\bctitle{Likelihood ratios for out-of-distribution detection}.
In: \bbtitle{Advances in Neural Information Processing Systems},
vol. \bseriesno{32}.
\bpublisher{Curran Associates, Inc.},
\blocation{Vancouver}
(\byear{2019})
\end{bchapter}
\endbibitem

\bibitem[\protect\citeauthoryear{Xiao et~al.}{2020}]{xiao2020likelihood}
\begin{bchapter}
\bauthor{\bsnm{Xiao}, \binits{Z.}},
\bauthor{\bsnm{Yan}, \binits{Q.}},
\bauthor{\bsnm{Amit}, \binits{Y.}}:
\bctitle{Likelihood regret: An out-of-distribution detection score for
  variational auto-encoder}.
In: \bbtitle{Advances in Neural Information Processing Systems},
vol. \bseriesno{33},
pp. \bfpage{20685}--\blpage{20696}.
\bpublisher{Curran Associates, Inc.},
\blocation{Virtual}
(\byear{2020})
\end{bchapter}
\endbibitem

\bibitem[\protect\citeauthoryear{Serr{\`{a}} et~al.}{2020}]{Serrà2020Input}
\begin{bchapter}
\bauthor{\bsnm{Serr{\`{a}}}, \binits{J.}},
\bauthor{\bsnm{{\'{A}}lvarez}, \binits{D.}},
\bauthor{\bsnm{G{\'{o}}mez}, \binits{V.}},
\bauthor{\bsnm{Slizovskaia}, \binits{O.}},
\bauthor{\bsnm{N{\'{u}}{\~{n}}ez}, \binits{J.F.}},
\bauthor{\bsnm{Luque}, \binits{J.}}:
\bctitle{Input complexity and out-of-distribution detection with
  likelihood-based generative models}.
In: \bbtitle{International Conference on Learning Representations}.
\bpublisher{OpenReview.net},
\blocation{Addis Ababa}
(\byear{2020})
\end{bchapter}
\endbibitem

\bibitem[\protect\citeauthoryear{Kingma and Welling}{2014}]{kingma2013auto}
\begin{bchapter}
\bauthor{\bsnm{Kingma}, \binits{D.P.}},
\bauthor{\bsnm{Welling}, \binits{M.}}:
\bctitle{Auto-encoding variational bayes}.
In: \bbtitle{International Conference on Learning Representations},
\bconflocation{Banff}
(\byear{2014})
\end{bchapter}
\endbibitem

\bibitem[\protect\citeauthoryear{Goodfellow
  et~al.}{2014}]{goodfellow2014generative}
\begin{bchapter}
\bauthor{\bsnm{Goodfellow}, \binits{I.}},
\bauthor{\bsnm{Pouget-Abadie}, \binits{J.}},
\bauthor{\bsnm{Mirza}, \binits{M.}},
\bauthor{\bsnm{Xu}, \binits{B.}},
\bauthor{\bsnm{Warde-Farley}, \binits{D.}},
\bauthor{\bsnm{Ozair}, \binits{S.}},
\bauthor{\bsnm{Courville}, \binits{A.}},
\bauthor{\bsnm{Bengio}, \binits{Y.}}:
\bctitle{Generative adversarial nets}.
In: \bbtitle{Advances in Neural Information Processing Systems},
vol. \bseriesno{27}.
\bpublisher{Curran Associates, Inc.},
\blocation{Montreal}
(\byear{2014})
\end{bchapter}
\endbibitem

\bibitem[\protect\citeauthoryear{Brehmer and Cranmer}{2020}]{brehmer2020flows}
\begin{bchapter}
\bauthor{\bsnm{Brehmer}, \binits{J.}},
\bauthor{\bsnm{Cranmer}, \binits{K.}}:
\bctitle{Flows for simultaneous manifold learning and density estimation}.
In: \bbtitle{Advances in Neural Information Processing Systems},
vol. \bseriesno{33},
pp. \bfpage{442}--\blpage{453}.
\bpublisher{Curran Associates, Inc.},
\blocation{Virtual}
(\byear{2020})
\end{bchapter}
\endbibitem

\bibitem[\protect\citeauthoryear{Caterini
  et~al.}{2021}]{caterini2021rectangular}
\begin{bchapter}
\bauthor{\bsnm{Caterini}, \binits{A.L.}},
\bauthor{\bsnm{Loaiza-Ganem}, \binits{G.}},
\bauthor{\bsnm{Pleiss}, \binits{G.}},
\bauthor{\bsnm{Cunningham}, \binits{J.P.}}:
\bctitle{Rectangular flows for manifold learning}.
In: \bbtitle{Advances in Neural Information Processing Systems},
vol. \bseriesno{34},
pp. \bfpage{30228}--\blpage{30241}.
\bpublisher{Curran Associates, Inc.},
\blocation{Virtual}
(\byear{2021})
\end{bchapter}
\endbibitem

\bibitem[\protect\citeauthoryear{Huang
  et~al.}{2021}]{cunningham2020normalizing}
\begin{bchapter}
\bauthor{\bsnm{Huang}, \binits{C.-W.}},
\bauthor{\bsnm{Krueger}, \binits{D.}},
\bauthor{\bsnm{Berg}, \binits{R.V.}},
\bauthor{\bsnm{Papamakarios}, \binits{G.}},
\bauthor{\bsnm{Cremer}, \binits{C.}},
\bauthor{\bsnm{Chen}, \binits{R.T.Q.}},
\bauthor{\bsnm{Rezende}, \binits{D.J.}}:
\bctitle{Normalizing flows across dimensions}.
In: \bbtitle{International Conference on Machine Learning}.
\bsertitle{Workshop on Invertible Neural Networks, Normalizing Flows, and
  Explicit Likelihood Models}
(\byear{2021})
\end{bchapter}
\endbibitem

\bibitem[\protect\citeauthoryear{Horvat and
  Pfister}{2021}]{horvat2021denoising}
\begin{bchapter}
\bauthor{\bsnm{Horvat}, \binits{C.}},
\bauthor{\bsnm{Pfister}, \binits{J.-P.}}:
\bctitle{Denoising normalizing flow}.
In: \bbtitle{Advances in Neural Information Processing Systems},
vol. \bseriesno{34},
pp. \bfpage{9099}--\blpage{9111}.
\bpublisher{Curran Associates, Inc.},
\blocation{Virtual}
(\byear{2021})
\end{bchapter}
\endbibitem

\bibitem[\protect\citeauthoryear{Huber}{1964}]{huber}
\begin{barticle}
\bauthor{\bsnm{Huber}, \binits{P.J.}}:
\batitle{Robust estimation of a location parameter}.
\bjtitle{The Annals of Mathematical Statistics}
\bvolume{35}(\bissue{1}),
\bfpage{73}--\blpage{101}
(\byear{1964})
\end{barticle}
\endbibitem

\bibitem[\protect\citeauthoryear{Kim et~al.}{2020}]{kim2020softflow}
\begin{bchapter}
\bauthor{\bsnm{Kim}, \binits{H.}},
\bauthor{\bsnm{Lee}, \binits{H.}},
\bauthor{\bsnm{Kang}, \binits{W.H.}},
\bauthor{\bsnm{Lee}, \binits{J.Y.}},
\bauthor{\bsnm{Kim}, \binits{N.S.}}:
\bctitle{{S}oft{F}low: Probabilistic framework for normalizing flow on
  manifolds}.
In: \bbtitle{Advances in Neural Information Processing Systems},
vol. \bseriesno{33},
pp. \bfpage{16388}--\blpage{16397}.
\bpublisher{Curran Associates, Inc.},
\blocation{Virtual}
(\byear{2020})
\end{bchapter}
\endbibitem

\bibitem[\protect\citeauthoryear{Kothari et~al.}{2021}]{kothari2021trumpets}
\begin{bchapter}
\bauthor{\bsnm{Kothari}, \binits{K.}},
\bauthor{\bsnm{Khorashadizadeh}, \binits{A.}},
\bauthor{\bsnm{Hoop}, \binits{M.V.}},
\bauthor{\bsnm{Dokmanic}, \binits{I.}}:
\bctitle{Trumpets: Injective flows for inference and inverse problems}.
In: \bbtitle{Conference on Uncertainty in Artificial Intelligence}.
\bsertitle{Proceedings of Machine Learning Research},
vol. \bseriesno{161},
pp. \bfpage{1269}--\blpage{1278}.
\bpublisher{{AUAI} Press},
\blocation{Virtual}
(\byear{2021})
\end{bchapter}
\endbibitem

\bibitem[\protect\citeauthoryear{Kalatzis et~al.}{2022}]{kalatzis2021multi}
\begin{botherref}
\oauthor{\bsnm{Kalatzis}, \binits{D.}},
\oauthor{\bsnm{Ye}, \binits{J.Z.}},
\oauthor{\bsnm{Pouplin}, \binits{A.}},
\oauthor{\bsnm{Wohlert}, \binits{J.}},
\oauthor{\bsnm{Hauberg}, \binits{S.}}:
Density estimation on smooth manifolds with normalizing flows.
Preprint at \url{https://arxiv.org/abs/2106.03500}
(2022)
\end{botherref}
\endbibitem

\bibitem[\protect\citeauthoryear{Ross and Cresswell}{2021}]{ross2021conformal}
\begin{bchapter}
\bauthor{\bsnm{Ross}, \binits{B.}},
\bauthor{\bsnm{Cresswell}, \binits{J.}}:
\bctitle{Tractable density estimation on learned manifolds with conformal
  embedding flows}.
In: \bbtitle{Advances in Neural Information Processing Systems},
vol. \bseriesno{34},
pp. \bfpage{26635}--\blpage{26648}.
\bpublisher{Curran Associates, Inc.},
\blocation{Virtual}
(\byear{2021})
\end{bchapter}
\endbibitem

\bibitem[\protect\citeauthoryear{Grcic et~al.}{2021}]{DBLP:conf/nips/GrcicGS21}
\begin{bchapter}
\bauthor{\bsnm{Grcic}, \binits{M.}},
\bauthor{\bsnm{Grubisic}, \binits{I.}},
\bauthor{\bsnm{Segvic}, \binits{S.}}:
\bctitle{Densely connected normalizing flows}.
In: \bbtitle{Advances in Neural Information Processing Systems},
vol. \bseriesno{34},
pp. \bfpage{23968}--\blpage{23982}.
\bpublisher{Curran Associates, Inc.},
\blocation{Virtual}
(\byear{2021})
\end{bchapter}
\endbibitem

\bibitem[\protect\citeauthoryear{Lee et~al.}{2018}]{lee2017training}
\begin{bchapter}
\bauthor{\bsnm{Lee}, \binits{K.}},
\bauthor{\bsnm{Lee}, \binits{H.}},
\bauthor{\bsnm{Lee}, \binits{K.}},
\bauthor{\bsnm{Shin}, \binits{J.}}:
\bctitle{Training confidence-calibrated classifiers for detecting
  out-of-distribution samples}.
In: \bbtitle{International Conference on Learning Representations}.
\bpublisher{OpenReview.net},
\blocation{Vancouver}
(\byear{2018})
\end{bchapter}
\endbibitem

\bibitem[\protect\citeauthoryear{Hendrycks et~al.}{2019}]{hendrycks2018deep}
\begin{bchapter}
\bauthor{\bsnm{Hendrycks}, \binits{D.}},
\bauthor{\bsnm{Mazeika}, \binits{M.}},
\bauthor{\bsnm{Dietterich}, \binits{T.G.}}:
\bctitle{Deep anomaly detection with outlier exposure}.
In: \bbtitle{International Conference on Learning Representations}.
\bpublisher{OpenReview.net},
\blocation{Los Angeles}
(\byear{2019})
\end{bchapter}
\endbibitem

\bibitem[\protect\citeauthoryear{DeVries and
  Taylor}{2018}]{devries2018learning}
\begin{botherref}
\oauthor{\bsnm{DeVries}, \binits{T.}},
\oauthor{\bsnm{Taylor}, \binits{G.W.}}:
Learning Confidence for Out-of-Distribution Detection in Neural Networks
(2018)
\end{botherref}
\endbibitem

\bibitem[\protect\citeauthoryear{Liang et~al.}{2018}]{liang2018enhancing}
\begin{bchapter}
\bauthor{\bsnm{Liang}, \binits{S.}},
\bauthor{\bsnm{Li}, \binits{Y.}},
\bauthor{\bsnm{Srikant}, \binits{R.}}:
\bctitle{Enhancing the reliability of out-of-distribution image detection in
  neural networks}.
In: \bbtitle{International Conference on Learning Representations}.
\bpublisher{OpenReview.net},
\blocation{Vancouver}
(\byear{2018})
\end{bchapter}
\endbibitem

\bibitem[\protect\citeauthoryear{Zhang et~al.}{2021}]{UnderstandingBackground}
\begin{bchapter}
\bauthor{\bsnm{Zhang}, \binits{L.H.}},
\bauthor{\bsnm{Goldstein}, \binits{M.}},
\bauthor{\bsnm{Ranganath}, \binits{R.}}:
\bctitle{Understanding failures in out-of-distribution detection with deep
  generative models}.
In: \bbtitle{International Conference on Machine Learning}.
\bsertitle{Proceedings of Machine Learning Research},
vol. \bseriesno{139},
pp. \bfpage{12427}--\blpage{12436}.
\bpublisher{{PMLR}},
\blocation{Virtual}
(\byear{2021})
\end{bchapter}
\endbibitem

\bibitem[\protect\citeauthoryear{Nagarajan et~al.}{2021}]{UnderstandingFailure}
\begin{bchapter}
\bauthor{\bsnm{Nagarajan}, \binits{V.}},
\bauthor{\bsnm{Andreassen}, \binits{A.}},
\bauthor{\bsnm{Neyshabur}, \binits{B.}}:
\bctitle{Understanding the failure modes of out-of-distribution
  generalization}.
In: \bbtitle{International Conference on Learning Representations}.
\bpublisher{OpenReview.net},
\blocation{Virtual}
(\byear{2021})
\end{bchapter}
\endbibitem

\bibitem[\protect\citeauthoryear{Kumar et~al.}{2021}]{kumar2021inflow}
\begin{botherref}
\oauthor{\bsnm{Kumar}, \binits{N.}},
\oauthor{\bsnm{Hanfeld}, \binits{P.}},
\oauthor{\bsnm{Hecht}, \binits{M.}},
\oauthor{\bsnm{Bussmann}, \binits{M.}},
\oauthor{\bsnm{Gumhold}, \binits{S.}},
\oauthor{\bsnm{Hoffmann}, \binits{N.}}:
InFlow: Robust outlier detection utilizing Normalizing Flows.
Preprint at \url{https://arxiv.org/abs/2106.12894}
(2021)
\end{botherref}
\endbibitem

\bibitem[\protect\citeauthoryear{Zisselman and Tamar}{2020}]{zisselman2020deep}
\begin{bchapter}
\bauthor{\bsnm{Zisselman}, \binits{E.}},
\bauthor{\bsnm{Tamar}, \binits{A.}}:
\bctitle{Deep residual flow for out of distribution detection}.
In: \bbtitle{{IEEE/CVF} Conference on Computer Vision and Pattern Recognition},
pp. \bfpage{13991}--\blpage{14000}.
\bpublisher{Computer Vision Foundation / {IEEE}},
\blocation{Seattle}
(\byear{2020})
\end{bchapter}
\endbibitem

\bibitem[\protect\citeauthoryear{Ardizzone
  et~al.}{2020}]{ardizzone2020training}
\begin{bchapter}
\bauthor{\bsnm{Ardizzone}, \binits{L.}},
\bauthor{\bsnm{Mackowiak}, \binits{R.}},
\bauthor{\bsnm{Rother}, \binits{C.}},
\bauthor{\bsnm{K\"{o}the}, \binits{U.}}:
\bctitle{Training normalizing flows with the information bottleneck for
  competitive generative classification}.
In: \bbtitle{Advances in Neural Information Processing Systems},
vol. \bseriesno{33},
pp. \bfpage{7828}--\blpage{7840}.
\bpublisher{Curran Associates, Inc.},
\blocation{Virtual}
(\byear{2020})
\end{bchapter}
\endbibitem

\bibitem[\protect\citeauthoryear{Choi et~al.}{2019}]{choi2018waic}
\begin{botherref}
\oauthor{\bsnm{Choi}, \binits{H.}},
\oauthor{\bsnm{Jang}, \binits{E.}},
\oauthor{\bsnm{Alemi}, \binits{A.A.}}:
WAIC, but Why? Generative Ensembles for Robust Anomaly Detection
(2019)
\end{botherref}
\endbibitem

\bibitem[\protect\citeauthoryear{Cho et~al.}{2022}]{LandR}
\begin{barticle}
\bauthor{\bsnm{Cho}, \binits{M.}},
\bauthor{\bsnm{Kim}, \binits{T.}},
\bauthor{\bsnm{Kim}, \binits{W.J.}},
\bauthor{\bsnm{Cho}, \binits{S.}},
\bauthor{\bsnm{Lee}, \binits{S.}}:
\batitle{Unsupervised video anomaly detection via normalizing flows with
  implicit latent features}.
\bjtitle{Pattern Recognit.}
\bvolume{129},
\bfpage{108703}
(\byear{2022})
\end{barticle}
\endbibitem

\bibitem[\protect\citeauthoryear{Zhao et~al.}{2023}]{medicalanomaly}
\begin{bchapter}
\bauthor{\bsnm{Zhao}, \binits{Y.}},
\bauthor{\bsnm{Ding}, \binits{Q.}},
\bauthor{\bsnm{Zhang}, \binits{X.}}:
\bctitle{{AE-FLOW:} autoencoders with normalizing flows for medical images
  anomaly detection}.
In: \bbtitle{International Conference on Learning Representations}.
\bpublisher{OpenReview.net},
\blocation{Kigali}
(\byear{2023})
\end{bchapter}
\endbibitem

\bibitem[\protect\citeauthoryear{Rezende and
  Mohamed}{2015}]{rezende2015variational}
\begin{bchapter}
\bauthor{\bsnm{Rezende}, \binits{D.J.}},
\bauthor{\bsnm{Mohamed}, \binits{S.}}:
\bctitle{Variational inference with normalizing flows}.
In: \bbtitle{International Conference on Machine Learning}.
\bsertitle{Workshop and Conference Proceedings},
vol. \bseriesno{37},
pp. \bfpage{1530}--\blpage{1538}.
\bpublisher{JMLR.org},
\blocation{Lille}
(\byear{2015})
\end{bchapter}
\endbibitem

\bibitem[\protect\citeauthoryear{Murphy}{2023}]{murphy2022probabilistic}
\begin{bbook}
\bauthor{\bsnm{Murphy}, \binits{K.P.}}:
\bbtitle{Probabilistic Machine Learning: Advanced Topics}.
\bpublisher{MIT Press}, \blocation{???}
(\byear{2023}).
\bcomment{Preprint at \url{http://probml.github.io/book2}}
\end{bbook}
\endbibitem

\bibitem[\protect\citeauthoryear{Ho et~al.}{2020}]{NEURIPS2020_4c5bcfec}
\begin{bchapter}
\bauthor{\bsnm{Ho}, \binits{J.}},
\bauthor{\bsnm{Jain}, \binits{A.}},
\bauthor{\bsnm{Abbeel}, \binits{P.}}:
\bctitle{Denoising diffusion probabilistic models}.
In: \bbtitle{Advances in Neural Information Processing Systems},
vol. \bseriesno{33},
pp. \bfpage{6840}--\blpage{6851}.
\bpublisher{Curran Associates, Inc.},
\blocation{Virtual}
(\byear{2020})
\end{bchapter}
\endbibitem

\bibitem[\protect\citeauthoryear{Graham
  et~al.}{2023}]{DBLP:conf/cvpr/GrahamPTNOC23}
\begin{bchapter}
\bauthor{\bsnm{Graham}, \binits{M.S.}},
\bauthor{\bsnm{Pinaya}, \binits{W.H.L.}},
\bauthor{\bsnm{Tudosiu}, \binits{P.}},
\bauthor{\bsnm{Nachev}, \binits{P.}},
\bauthor{\bsnm{Ourselin}, \binits{S.}},
\bauthor{\bsnm{Cardoso}, \binits{M.J.}}:
\bctitle{Denoising diffusion models for out-of-distribution detection}.
In: \bbtitle{Conference on Computer Vision and Pattern Recognition},
pp. \bfpage{2948}--\blpage{2957}.
\bpublisher{{IEEE}},
\blocation{Vancouver, BC, Canada}
(\byear{2023})
\end{bchapter}
\endbibitem

\bibitem[\protect\citeauthoryear{LeCun and Cortes}{2010}]{mnist}
\begin{botherref}
\oauthor{\bsnm{LeCun}, \binits{Y.}},
\oauthor{\bsnm{Cortes}, \binits{C.}}:
The {MNIST} database of handwritten digits.
\url{http://yann.lecun.com/exdb/mnist}
(2010)
\end{botherref}
\endbibitem

\bibitem[\protect\citeauthoryear{Xiao et~al.}{2017}]{xiao2017fashionmnist}
\begin{botherref}
\oauthor{\bsnm{Xiao}, \binits{H.}},
\oauthor{\bsnm{Rasul}, \binits{K.}},
\oauthor{\bsnm{Vollgraf}, \binits{R.}}:
Fashion-{MNIST}: a Novel Image Dataset for Benchmarking Machine Learning
  Algorithms.
Preprint at \url{https://arxiv.org/abs/1708.07747}
(2017)
\end{botherref}
\endbibitem

\bibitem[\protect\citeauthoryear{Netzer et~al.}{2011}]{svhn}
\begin{bchapter}
\bauthor{\bsnm{Netzer}, \binits{Y.}},
\bauthor{\bsnm{Wang}, \binits{T.}},
\bauthor{\bsnm{Coates}, \binits{A.}},
\bauthor{\bsnm{Bissacco}, \binits{A.}},
\bauthor{\bsnm{Wu}, \binits{B.}},
\bauthor{\bsnm{Ng}, \binits{A.Y.}}:
\bctitle{Reading digits in natural images with unsupervised feature learning}.
In: \bbtitle{Advances in Neural Information Processing Systems}.
\bsertitle{Workshop on Deep Learning and Unsupervised Feature Learning},
\bconflocation{Granada}
(\byear{2011}).
\burl{http://ufldl.stanford.edu/housenumbers/}
\end{bchapter}
\endbibitem

\bibitem[\protect\citeauthoryear{Krizhevsky et~al.}{2009}]{cifar10}
\begin{botherref}
\oauthor{\bsnm{Krizhevsky}, \binits{A.}},
\oauthor{\bsnm{Nair}, \binits{V.}},
\oauthor{\bsnm{Hinton}, \binits{G.}}:
The {CIFAR}-10 dataset.
\url{https://www.cs.toronto.edu/~kriz/cifar.html}
(2009)
\end{botherref}
\endbibitem

\bibitem[\protect\citeauthoryear{Liu et~al.}{2015}]{liu2015faceattributes}
\begin{bchapter}
\bauthor{\bsnm{Liu}, \binits{Z.}},
\bauthor{\bsnm{Luo}, \binits{P.}},
\bauthor{\bsnm{Wang}, \binits{X.}},
\bauthor{\bsnm{Tang}, \binits{X.}}:
\bctitle{Deep learning face attributes in the wild}.
In: \bbtitle{{IEEE} International Conference on Computer Vision},
pp. \bfpage{3730}--\blpage{3738}.
\bpublisher{{IEEE} Computer Society},
\blocation{Santiago}
(\byear{2015})
\end{bchapter}
\endbibitem

\bibitem[\protect\citeauthoryear{Yu et~al.}{2016}]{yu2015lsun}
\begin{botherref}
\oauthor{\bsnm{Yu}, \binits{F.}},
\oauthor{\bsnm{Seff}, \binits{A.}},
\oauthor{\bsnm{Zhang}, \binits{Y.}},
\oauthor{\bsnm{Song}, \binits{S.}},
\oauthor{\bsnm{Funkhouser}, \binits{T.}},
\oauthor{\bsnm{Xiao}, \binits{J.}}:
LSUN: Construction of a Large-scale Image Dataset using Deep Learning with
  Humans in the Loop.
Preprint at \url{https://arxiv.org/abs/1506.03365}
(2016)
\end{botherref}
\endbibitem

\bibitem[\protect\citeauthoryear{Kingma and Dhariwal}{2018}]{kingma2018glow}
\begin{bchapter}
\bauthor{\bsnm{Kingma}, \binits{D.P.}},
\bauthor{\bsnm{Dhariwal}, \binits{P.}}:
\bctitle{{G}low: Generative flow with invertible 1x1 convolutions}.
In: \bbtitle{Advances in Neural Information Processing Systems},
vol. \bseriesno{31}.
\bpublisher{Curran Associates, Inc.},
\blocation{Montreal}
(\byear{2018})
\end{bchapter}
\endbibitem

\bibitem[\protect\citeauthoryear{Dinh et~al.}{2017}]{dinh2016density}
\begin{bchapter}
\bauthor{\bsnm{Dinh}, \binits{L.}},
\bauthor{\bsnm{Sohl{-}Dickstein}, \binits{J.}},
\bauthor{\bsnm{Bengio}, \binits{S.}}:
\bctitle{Density estimation using {Real NVP}}.
In: \bbtitle{International Conference on Learning Representations}.
\bpublisher{OpenReview.net},
\blocation{Toulon}
(\byear{2017})
\end{bchapter}
\endbibitem

\bibitem[\protect\citeauthoryear{Kingma and Ba}{2015}]{adam}
\begin{bchapter}
\bauthor{\bsnm{Kingma}, \binits{D.P.}},
\bauthor{\bsnm{Ba}, \binits{J.}}:
\bctitle{Adam: a method for stochastic optimization}.
In: \bbtitle{International Conference on Learning Representations},
\bconflocation{San Diego}
(\byear{2015})
\end{bchapter}
\endbibitem

\bibitem[\protect\citeauthoryear{Horvat and
  Pfister}{2022}]{NEURIPS2022_4f918fa3}
\begin{bchapter}
\bauthor{\bsnm{Horvat}, \binits{C.}},
\bauthor{\bsnm{Pfister}, \binits{J.-P.}}:
\bctitle{Intrinsic dimensionality estimation using normalizing flows}.
In: \beditor{\bsnm{Koyejo}, \binits{S.}},
\beditor{\bsnm{Mohamed}, \binits{S.}},
\beditor{\bsnm{Agarwal}, \binits{A.}},
\beditor{\bsnm{Belgrave}, \binits{D.}},
\beditor{\bsnm{Cho}, \binits{K.}},
\beditor{\bsnm{Oh}, \binits{A.}} (eds.)
\bbtitle{Advances in Neural Information Processing Systems},
vol. \bseriesno{35},
pp. \bfpage{12225}--\blpage{12236}.
\bpublisher{Curran Associates, Inc.},
\blocation{Los Angeles}
(\byear{2022})
\end{bchapter}
\endbibitem

\bibitem[\protect\citeauthoryear{Pope et~al.}{2021}]{DBLP:conf/iclr/PopeZAGG21}
\begin{bchapter}
\bauthor{\bsnm{Pope}, \binits{P.}},
\bauthor{\bsnm{Zhu}, \binits{C.}},
\bauthor{\bsnm{Abdelkader}, \binits{A.}},
\bauthor{\bsnm{Goldblum}, \binits{M.}},
\bauthor{\bsnm{Goldstein}, \binits{T.}}:
\bctitle{The intrinsic dimension of images and its impact on learning}.
In: \bbtitle{International Conference on Learning Representations}.
\bpublisher{OpenReview.net},
\blocation{Virtual}
(\byear{2021})
\end{bchapter}
\endbibitem

\bibitem[\protect\citeauthoryear{Karras
  et~al.}{2020}]{DBLP:conf/cvpr/KarrasLAHLA20}
\begin{bchapter}
\bauthor{\bsnm{Karras}, \binits{T.}},
\bauthor{\bsnm{Laine}, \binits{S.}},
\bauthor{\bsnm{Aittala}, \binits{M.}},
\bauthor{\bsnm{Hellsten}, \binits{J.}},
\bauthor{\bsnm{Lehtinen}, \binits{J.}},
\bauthor{\bsnm{Aila}, \binits{T.}}:
\bctitle{Analyzing and improving the image quality of stylegan}.
In: \bbtitle{{IEEE/CVF} Conference on Computer Vision and Pattern Recognition},
pp. \bfpage{8107}--\blpage{8116}.
\bpublisher{Computer Vision Foundation / IEEE},
\blocation{Seattle}
(\byear{2020})
\end{bchapter}
\endbibitem

\end{thebibliography}

\end{document}


\title[Article Title]{Supplementary Materials for: Out-of-distribution detection using normalizing flows on the data manifold}

%
%
%
%


\maketitle

\section{Image generation}
\label{appendix:image_generation}

Due to page limitations in the paper, manifold learning experiments on some other datasets (CelebA, SVHN) are provided here.

\subsection{Cifar10}

\begin{figure}[h]
\centering
\subfigure[$\mathcal{M}_G \subset \mathbb{R}^{10}$]{\includegraphics[width=0.18\textwidth]{\FigPath{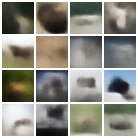}}}
\subfigure[$\mathcal{M}_G \subset \mathbb{R}^{50}$]{\includegraphics[width=0.18\textwidth]{\FigPath{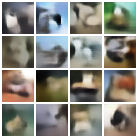}}}
\subfigure[$\mathcal{M}_G \subset \mathbb{R}^{100}$]{\includegraphics[width=0.18\textwidth]{\FigPath{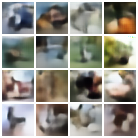}}}
\subfigure[$\mathcal{M}_G \subset \mathbb{R}^{500}$]{\includegraphics[width=0.18\textwidth]{\FigPath{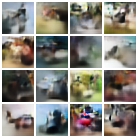}}}
\subfigure[$\mathcal{M}_G \subset \mathbb{R}^{1000}$]{\includegraphics[width=0.18\textwidth]{\FigPath{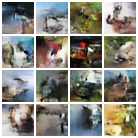}}}
\\
\subfigure[$\mathcal{M}_R \subset \mathbb{R}^{10}$]{\includegraphics[width=0.18\textwidth]{\FigPath{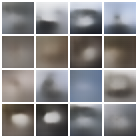}}}
\subfigure[$\mathcal{M}_R \subset \mathbb{R}^{50}$]{\includegraphics[width=0.18\textwidth]{\FigPath{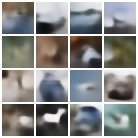}}}
\subfigure[$\mathcal{M}_R \subset \mathbb{R}^{100}$]{\includegraphics[width=0.18\textwidth]{\FigPath{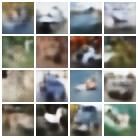}}}
\subfigure[$\mathcal{M}_R \subset \mathbb{R}^{500}$]{\includegraphics[width=0.18\textwidth]{\FigPath{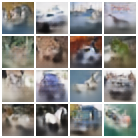}}}
\subfigure[$\mathcal{M}_R \subset \mathbb{R}^{1000}$]{\includegraphics[width=0.18\textwidth]{\FigPath{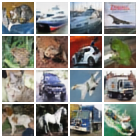}}}
\\
\subfigure[$\mathcal{M}_R \subset \mathbb{R}^{10}$]{\includegraphics[width=0.18\textwidth]{\FigPath{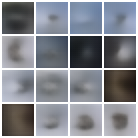}}}
\subfigure[$\mathcal{M}_R \subset \mathbb{R}^{50}$]{\includegraphics[width=0.18\textwidth]{\FigPath{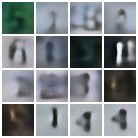}}}
\subfigure[$\mathcal{M}_R \subset \mathbb{R}^{100}$]{\includegraphics[width=0.18\textwidth]{\FigPath{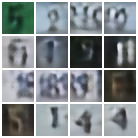}}}
\subfigure[$\mathcal{M}_R \subset \mathbb{R}^{500}$]{\includegraphics[width=0.18\textwidth]{\FigPath{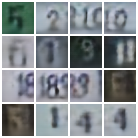}}}
\subfigure[$\mathcal{M}_R \subset \mathbb{R}^{1000}$]{\includegraphics[width=0.18\textwidth]{\FigPath{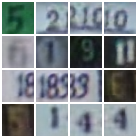}}}
\\
\subfigure[$\mathcal{M}_R \subset \mathbb{R}^{10}$]{\includegraphics[width=0.18\textwidth]{\FigPath{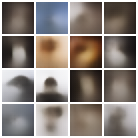}}}
\subfigure[$\mathcal{M}_R \subset \mathbb{R}^{50}$]{\includegraphics[width=0.18\textwidth]{\FigPath{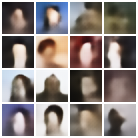}}}
\subfigure[$\mathcal{M}_R \subset \mathbb{R}^{100}$]{\includegraphics[width=0.18\textwidth]{\FigPath{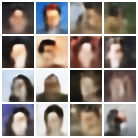}}}
\subfigure[$\mathcal{M}_R \subset \mathbb{R}^{500}$]{\includegraphics[width=0.18\textwidth]{\FigPath{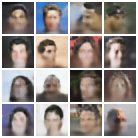}}}
\subfigure[$\mathcal{M}_R \subset \mathbb{R}^{1000}$]{\includegraphics[width=0.18\textwidth]{\FigPath{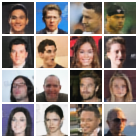}}}

\caption{
Several generated (first row) and reconstructed images (Cifar10 as ID data in the second row, SVHN and CelebA as OOD data in the third and forth rows, respectively) from proposed-D method for different manifold dimensions (10, 50, 100, 500, and 1000 from left to right) for a model trained on the Cifar10 dataset.
$\mathcal{M}_G \subset \mathbb{R}^{d}$ and $\mathcal{M}_R \subset \mathbb{R}^{d}$ represent image generation and image reconstruction from a manifold of dimension $d$, respectively.
}
\label{fig:figure_Cifar10_generation_reconstruction}
\end{figure}

A same performance to CelebA (reported in the main paper) can also be observed for a trained model on Cifar10 in \FigRef{fig:figure_Cifar10_generation_reconstruction}.
As it is known, the generated data (first row) is so blurry for low dimensions, and only the overall structure of objects is distinguishable.
However, increasing dimensions leads to collapsing the border of objects due to learning more details.
In cases where the data manifold is visually simple (such as reported results for SVHN in the next part), the likelihood is fitted better, and so sharpened images are generated with dimension increasing.
The second row of \FigRef{fig:figure_Cifar10_generation_reconstruction} contains the reconstruction of Cifar10 test data on the model trained with Cifar10 for the corresponding manifold dimension. The reconstruction ability is improved by increasing the dimension.
As it is clear, the model's invertibility strength will appear by increasing manifold dimension, the same as standard NF.
Furthermore, the ability of the model to learn the manifold in face of OOD data is demonstrated in the third and fourth rows in \FigRef{fig:figure_Cifar10_generation_reconstruction}. As evident from the results, lower dimensions preserve the essential information of the ID manifold.

\subsection{SVHN}

\begin{figure}[h]
\centering

\subfigure[$\mathcal{M}_G \subset \mathbb{R}^{10}$]{\includegraphics[width=0.18\textwidth]{\FigPath{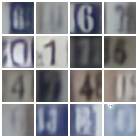}}}
\subfigure[$\mathcal{M}_G \subset \mathbb{R}^{50}$]{\includegraphics[width=0.18\textwidth]{\FigPath{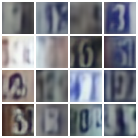}}}
\subfigure[$\mathcal{M}_G \subset \mathbb{R}^{100}$]{\includegraphics[width=0.18\textwidth]{\FigPath{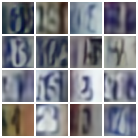}}}
\subfigure[$\mathcal{M}_G \subset \mathbb{R}^{500}$]{\includegraphics[width=0.18\textwidth]{\FigPath{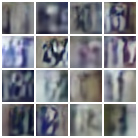}}}
\subfigure[$\mathcal{M}_G \subset \mathbb{R}^{1000}$]{\includegraphics[width=0.18\textwidth]{\FigPath{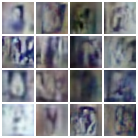}}}
\\
\subfigure[$\mathcal{M}_R \subset \mathbb{R}^{10}$]{\includegraphics[width=0.18\textwidth]{\FigPath{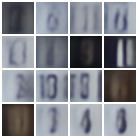}}}
\subfigure[$\mathcal{M}_R \subset \mathbb{R}^{50}$]{\includegraphics[width=0.18\textwidth]{\FigPath{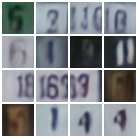}}}
\subfigure[$\mathcal{M}_R \subset \mathbb{R}^{100}$]{\includegraphics[width=0.18\textwidth]{\FigPath{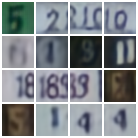}}}
\subfigure[$\mathcal{M}_R \subset \mathbb{R}^{500}$]{\includegraphics[width=0.18\textwidth]{\FigPath{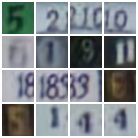}}}
\subfigure[$\mathcal{M}_R \subset \mathbb{R}^{1000}$]{\includegraphics[width=0.18\textwidth]{\FigPath{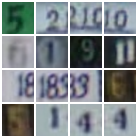}}}
\\
\subfigure[$\mathcal{M}_R \subset \mathbb{R}^{10}$]{\includegraphics[width=0.18\textwidth]{\FigPath{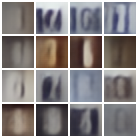}}}
\subfigure[$\mathcal{M}_R \subset \mathbb{R}^{50}$]{\includegraphics[width=0.18\textwidth]{\FigPath{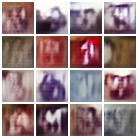}}}
\subfigure[$\mathcal{M}_R \subset \mathbb{R}^{100}$]{\includegraphics[width=0.18\textwidth]{\FigPath{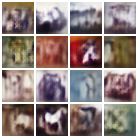}}}
\subfigure[$\mathcal{M}_R \subset \mathbb{R}^{500}$]{\includegraphics[width=0.18\textwidth]{\FigPath{learned_manifold_SVHN_500_reconstruction_cifar10.png}}}
\subfigure[$\mathcal{M}_R \subset \mathbb{R}^{1000}$]{\includegraphics[width=0.18\textwidth]{\FigPath{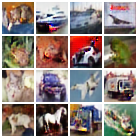}}}
\\
\subfigure[$\mathcal{M}_R \subset \mathbb{R}^{10}$]{\includegraphics[width=0.18\textwidth]{\FigPath{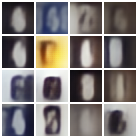}}}
\subfigure[$\mathcal{M}_R \subset \mathbb{R}^{50}$]{\includegraphics[width=0.18\textwidth]{\FigPath{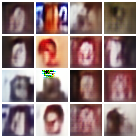}}}
\subfigure[$\mathcal{M}_R \subset \mathbb{R}^{100}$]{\includegraphics[width=0.18\textwidth]{\FigPath{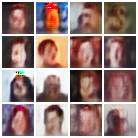}}}
\subfigure[$\mathcal{M}_R \subset \mathbb{R}^{500}$]{\includegraphics[width=0.18\textwidth]{\FigPath{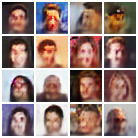}}}
\subfigure[$\mathcal{M}_R \subset \mathbb{R}^{1000}$]{\includegraphics[width=0.18\textwidth]{\FigPath{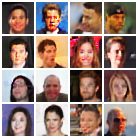}}}

\caption{
Several generated (first row) and reconstructed images (SVHN as ID data in the second row, Cifar10 and CelebA as OOD data in the third and forth rows, respectively)
from proposed-D method for different manifold dimensions (10, 50, 100, 500, and 1000 from left to right) for a model trained on the SVHN dataset.
$\mathcal{M}_G \subset \mathbb{R}^{d}$ and $\mathcal{M}_R \subset \mathbb{R}^{d}$ represent image generation and image reconstruction from a manifold fall in dimension $d$, respectively.
}
\label{fig:figure_SVHN_generation_reconstruction}
\end{figure}

Doing the same experiment to previouse section for SVHN as ID data is reported in \FigRef{fig:figure_SVHN_generation_reconstruction}.
As expected, the simplicity of SVHN manifold same as CelebA (reported in the main paper), leads to good performance on generation and reconstruction for ID data. While the learned manifold distribution for OOD data is not very suitable.
According to the NFs' literature, dimension-preserving NFs have error-less reconstruction. Experiments show that low-dimensional manifold learning (especially low dimensions) causes the model not to reconstruct OOD data perfectly.

\section{Architecture searching}
\label{appendix:architecture_searching_LSUN_dataset}

\begin{figure}[h]
\centering
\subfigure[Proposed-M, $\mathcal{M} \subset \mathbb{R}^{1000}$]{\includegraphics[width=0.30\textwidth]{\FigPath{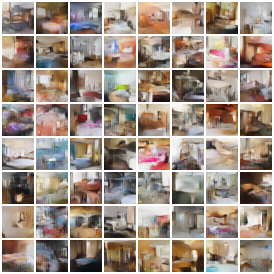}}}
\subfigure[Proposed-H, $\mathcal{M} \subset \mathbb{R}^{1000}$]{\includegraphics[width=0.30\textwidth]{\FigPath{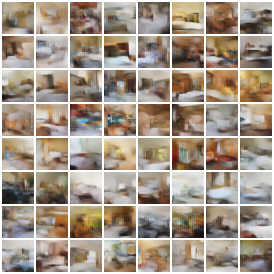}}}
\subfigure[DNF, $\mathcal{M} \subset \mathbb{R}^{1000}$]{\includegraphics[width=0.30\textwidth]{\FigPath{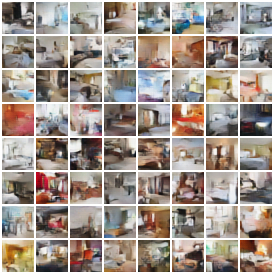}}}
\caption{
The randomly generated images for corresponding reported experiments in \TableRef{tab:table_lsun_architecture_searching}.
}
\label{fig:figure_lsun_architecture_searching}
\end{figure}

\begin{figure}[h]
\centering

\subfigure[Proposed-M, $\mathcal{M} \subset \mathbb{R}^{700}$]{\includegraphics[width=0.30\textwidth]{\FigPath{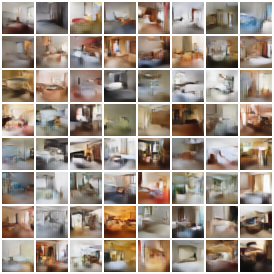}}}
\subfigure[Proposed-H, $\mathcal{M} \subset \mathbb{R}^{700}$]{\includegraphics[width=0.30\textwidth]{\FigPath{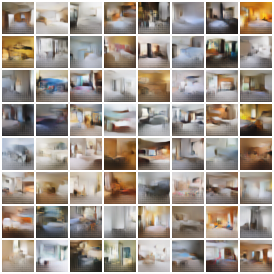}}}
\subfigure[DNF, $\mathcal{M} \subset \mathbb{R}^{700}$]{\includegraphics[width=0.30\textwidth]{\FigPath{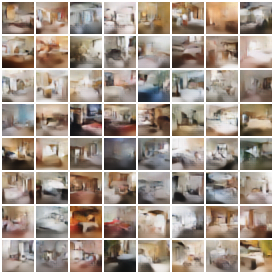}}}

\caption{
The randomly generated images for corresponding reported experiments in \TableRef{tab:figure_reduce_dimension_LSUN_dataset}.
}
\label{fig:figure_reduce_dimension_LSUN_dataset}
\end{figure}

Reported results about architecture searching in the main paper is included a well-defined face manifold dataset. However, it is also necessary to evaluate the model's architecture in crowded and cluttered manifolds. 
The class of bedrooms in the LSUN dataset is the next choice for doing experiments.
The corresponding results for the LSUN dataset for different configurations such as hierarchy in structure (Proposed-M, Proposed-H), and DNF are presented in \TableRef{tab:table_lsun_architecture_searching} and \FigRef{fig:figure_lsun_architecture_searching}.
The target manifold in the mentioneds experiment is $\mathbb{R}^{1000}$.
However, it seems possible to reach a manifold with a smaller dimension.
The results of next experiment are presented in \TableRef{tab:figure_reduce_dimension_LSUN_dataset} and \FigRef{fig:figure_reduce_dimension_LSUN_dataset} for an embedded manifold in $\mathbb{R}^{700}$.
The experiments confirm that the proposed framework, despite its structural similarity to DNF, leads to relatively better results in terms of numerical and visual results in the case of lower dimensions.

\begin{table}[h]
\setlength{\tabcolsep}{\tabcolsep}
\caption{The best MSE/BPD scores for the proposed methods and DNF on the LSUN dataset for target manifold embedded in $\mathbb{R}^{1000}$.}
\label{tab:table_lsun_architecture_searching}
\centering
\begin{tabularx}{\textwidth}{bbbb}
		\toprule
		
        \small Criterion
		& \makecell{\small Proposed-M \\ {\tiny ($3072 \rightarrow 2000 \rightarrow 1000$)}}
		& \makecell{\small Proposed-H \\ {\tiny ($3072 \rightarrow 2000 \rightarrow 1000$)}}
		& \makecell{\small DNF \\ {\tiny ($3072 \rightarrow 1000 \rightarrow 1000$)}}
		\\
		
		\cmidrule{1-4}\morecmidrules\cmidrule{1-4}
		\small MSE & \small 0.0003 $\pm$ 9e-6 & \small 0.0008 $\pm$ 2e-5 & \small 0.001 $\pm$ 9e-5 \\
		\small BPD & \small 3.65 $\pm$ 0.05 & \small 3.7 $\pm$ 0.05 & \small 4.10 $\pm$ 0.04 \\

		\bottomrule
\end{tabularx}
\end{table}

\begin{table}[h]
\caption{The best MSE/BPD scores for the proposed method and DNF on the LSUN dataset for target manifold embedded in $\mathbb{R}^{700}$.}
\label{tab:figure_reduce_dimension_LSUN_dataset}
\setlength{\tabcolsep}{\tabcolsep}
\centering
\begin{tabularx}{\textwidth}{bbbb}
		\toprule
		
		criterion
		& \makecell{Proposed-M \\ {\tiny ($3072 \rightarrow 1000 \rightarrow 700$)}}
		& \makecell{Proposed-H \\ {\tiny ($3072 \rightarrow 1000 \rightarrow 700$)}}
		& \makecell{DNF \\ {\tiny ($3072 \rightarrow 700 \rightarrow 700$)}}
		\\
		
		\cmidrule{1-4}\morecmidrules\cmidrule{1-4}
		MSE & 0.0004 $\pm$ 1e-5 & 0.0009 $\pm$ 2e-5 & 0.003 $\pm$ 0.0002 \\
		BPD & 3.66 $\pm$ 0.06 & 3.62 $\pm$ 0.06 & 4.14 $\pm$ 0.04 \\

		\bottomrule
\end{tabularx}
\end{table}

